\documentclass[10pt,letterpaper,oneside,final]{article}
\usepackage[T1]{fontenc}
\usepackage{subfigure,gensymb,amssymb,graphicx,booktabs}
\usepackage[ruled,vlined,linesnumbered]{algorithm2e}
\usepackage{textcomp,latexsym,tikz}
\usepackage{setspace}
\usepackage[font=small,labelfont=bf]{caption}
\usepackage{titling,authblk}
\usepackage{palatino}
\usepackage{fancyhdr}
\usepackage{hyperref}
\usepackage[margin=1.5in]{geometry}
\usetikzlibrary{matrix}
\usetikzlibrary{decorations.pathreplacing}
\graphicspath{{figs/}}
\providecommand{\keywords}[1]{\textbf{\textit{Keywords---}} #1}
\makeatletter
\renewcommand\@biblabel[1]{#1.}
\makeatother
\title{On Design Mining:\\ Coevolution and Surrogate Models}
\author{Richard J. Preen\thanks{Contact author. E-mail:~\url{richard2.preen@uwe.ac.uk}} }
\author{Larry Bull}
\affil{Department of Computer Science and Creative Technologies \\ University of the West of England, Bristol, UK}

\begin{document}
\maketitle
%\doublespacing

\begin{abstract}
Design mining is the use of computational intelligence techniques to iteratively search and model the attribute space of physical objects evaluated directly through rapid prototyping to meet given objectives. It enables the exploitation of novel materials and processes without formal models or complex simulation. In this paper, we focus upon the coevolutionary nature of the design process when it is decomposed into concurrent sub-design threads due to the overall complexity of the task. Using an abstract, tuneable model of coevolution we consider strategies to sample sub-thread designs for whole system testing and how best to construct and use surrogate models within the coevolutionary scenario. Drawing on our findings, the paper then describes the effective design of an array of six heterogeneous vertical-axis wind turbines. 
\end{abstract}


\clearpage
%\linenumbers
\setlength\tabcolsep{6pt}

\section{Introduction}

Design mining~\cite{Preen:2014,Preen:2015,Preen:2016} is the use of computational intelligence techniques to iteratively search and model the attribute space of physical objects evaluated directly through rapid prototyping to meet given objectives. It enables the exploitation of novel materials and processes without formal models or complex simulation, whilst harnessing the creativity of both computational and human design methods. A sample-model-search-sample loop creates an agile/flexible approach, i.e., primarily test-driven, enabling a continuing process of prototype design consideration and criteria refinement by both producers and users.  

Computational intelligence techniques have long been used in design, particularly for optimisation within simulations/models. Recent developments in additive-layer manufacturing (3D printing) means that it is now possible to work with over a hundred different materials, from ceramics to cells. In the simplest case, design mining assumes no prior knowledge and builds an initial model of the design space through the testing of 3D printed designs, whether specified by human and/or machine. Optimisation techniques, such as evolutionary algorithms (EAs), are then used to find the optima within the data mining model of the collected data; the model which maps design specifications to performance is inverted and suggested good solutions identified. These are then 3D printed and tested. The resulting data are added to the existing data and the process repeated. Over time the model---built solely from physical prototypes tested appropriately for the task requirements---captures the salient features of the design space, thereby enabling the discovery of high-quality (novel) solutions. Such so-called surrogate models have also long been used in optimisation for cases when simulations are computationally expensive. Their use with 3D printing opens new ways to exploit optimisation in the design of physical objects directly, whilst raising a number of new issues over the simulation case.  

This approach of constantly producing working prototypes from the beginning of the design process shares resemblance to agile software engineering~\cite{Martin:1991}: requirements are identified at the start, even if only partially, and then corresponding tests created, which are then used to drive the design process via rapid iterations of solution creation and evaluation. The constant supply of (tangible) prototypes enables informed sharing with, and hence feedback from, those involved in other stages of the process, such as those in manufacture or the end user. This feedback enables constant refinement of the requirements/testing and also means that aspects of the conceptual and detailed design stages become blended. Moreover, due to the constant production of better (physical) prototypes, aspects of the traditional manufacturing stage become merged with the design phase. The data mining models created provide new sources of knowledge, enabling designers, manufacturers, or users, to do what-if tests during the design process to suggest solutions, the sharing of timely/accurate information when concurrent sub-design threads are being exploited, etc. Thereafter, they serve as sources of information for further adaptive designs, the construction of simulators/models, etc. 

In contrast to human designers, who typically arrive at solutions by refining building blocks that have been identified in highly constrained ways, computational intelligence offers a much more unconstrained and unbiased approach  to exploring potential solutions. Thus, by creating the designs directly in hardware there is the potential that complex and subtle physical interactions can be utilised in unexpected ways where the operational principles were previously unknown. These physical effects may simply be insufficiently understood or absent from a simulator and thus otherwise unable to be exploited. Design mining is therefore ideally suited to applications involving highly complex environments and/or materials. 

The design of modern wind farms typically begin with the blade profile optimisation of a single isolated wind turbine through the use of computational fluid dynamics (CFD) simulations~\cite{Wang:2009}, followed by optimising the site positioning of multiple copies of the same design to minimise the negative effects of inter-turbine wake interactions~\cite{Gonzalez:2014}. Whilst CFD simulations have been successfully applied, they are extremely computationally expensive; consequently most numerical studies perform only 2D analysis, e.g., \cite{Ferreira:2015}, and it is currently infeasible to perform accurate 3D simulations of a large array. Moreover, various assumptions must be made, and accurately modelling the complex inter-turbine wake interactions is an extremely challenging task where different turbulence models can have a dramatic effect on turbine performance~\cite{Islam:2008}. CFD studies have also presented significant differences between results even with identical geometric and flow conditions due to the complexity of performing accurate numerical analysis~\cite{Akwa:2012}.

In our initial pilot study we used the design mining approach to discover a pair of novel, heterogeneous vertical-axis wind turbine (VAWT) designs through cooperative coevolution~\cite{Preen:2015}. Accurate and computationally efficient modelling of the inter-turbine interactions is extremely difficult and therefore the area is ideally suited to the design mining approach. More recently, we have begun to explore the performance of relevant techniques from the literature within the context of design mining. Following~\cite{Bull:1997a}, the pilot study used multi-layered perceptrons (MLPs)~\cite{Rosenblatt:1962} for the surrogate modelling. Using the data from that study, we have subsequently shown that MLPs appear a robust approach in comparison to a number of well-known techniques~\cite{Preen:2016}. That is, MLPs appear efficient at capturing the underlying structure of a design space from the relatively small amount of data points a physical sampling process can be expected to generate. In this paper we begin by continuing the line of enquiry, here focusing upon the coevolutionary nature of the design process when it is decomposed into concurrent sub-design threads due to the overall complexity of the task. Using an abstract, tuneable model of coevolution we consider strategies to sample sub-thread designs for whole system testing and how best to construct and use surrogate models within the coevolutionary scenario. Drawing on our findings, the paper then describes the effective design of a more complex array of VAWT than our pilot study. 

\section{Background}

\subsection{Evolving Physical Systems}

As we have reviewed elsewhere~\cite{Preen:2015}, there is a small amount of work considering the evolutionary design of physical systems directly, stretching back to the origins of the discipline~\cite{Box:1957,Pask:1958,Dunham:1963,Rechenberg:1971}. Well-known examples include robot controller design~\cite{Nolfi:1992}; the evolution of vertebrate tail stiffness in swimming robots~\cite{Long:2006}; adaptive antenna arrays~\cite{Becker:2014}; electronic circuit design using programmable hardware~\cite{Thompson:1998}; product design via human provided fitness values~\cite{Herdy:1996}; chemical systems~\cite{Theis:2006}; unconventional computers~\cite{Harding:2004}; robot embodied evolution~\cite{Ficici:1999}; drug discovery~\cite{Singh:1996}; functional genomics~\cite{King:2004}; adaptive optics~\cite{Sherman:2002}; quantum control~\cite{Judson:1992}; fermentation optimisation~\cite{Davies:2000}; and the optimisation of analytical instrumentation~\cite{OHagan:2005}. A selection of multiobjective case studies can be found in~\cite{Knowles:2009}. Examples of EAs incorporating the physical aerodynamic testing of candidate solutions include the optimisation of jet nozzles~\cite{Rechenberg:1971,Schwefel:1975}, as well as flapping~\cite{Augustsson:2002,Hunt:2005,Olhofer:2011} and morphing~\cite{Boria:2009} wings. More recent fluid dynamics examples include~\cite{Parezanovic:2015,Gautier:2015,Benard:2016}.

Lipson and Pollack~\cite{Lipson:2000} were the first to exploit the use of 3D printing in conjunction with an EA, printing mobile robots with embodied neural controllers that were evolved using a simulation of the mechanics and control. Rieffel and Sayles'~\cite{Rieffel:2010} use of an interactive EA to 3D print simple shapes is particularly relevant to the work presented here. As noted above, in this paper we adopt an approach where relatively simple and tuneable simulations of the basic evolutionary design scenario are used to explore the general performance of different algorithmic approaches before moving to the physical system. This can be seen as somewhat akin to the minimalist approach proposed by Jakobi~\cite{Jakobi:1997} for addressing the so-called reality gap in evolutionary robotics, although further removed from the full details of the physical system. Surrogate models are then used to capture the underlying characteristics of the system to guide design.

\subsection{Cooperative Coevolution and Surrogates}

Cooperative coevolution decomposes a global task into multiple interacting sub-component populations and optimises each in parallel. In the context of a design process, this can be seen as directly analogous to the use of concurrent sub-threads. The first known use of cooperative coevolution considered a job-shop scheduling task~\cite{Husbands:1991}. Here solutions for individual machines were first evaluated using a local fitness function before being partnered with solutions of equal rank in the other populations to create a global solution for evaluation. Bull and Fogarty~\cite{Bull:1993} subsequently presented a more general approach wherein the corresponding newly created offspring solutions from each population are partnered and evaluated. Later, Potter and De~Jong~\cite{Potter:1994} introduced a round-robin approach with each population evolved in turn, which has been adopted widely. They explored using the current best individual from each of the other populations to create a global solution, before extending it to using the best and a random individual from the other population(s). These two partnering strategies, along with others, were compared under their round-robin approach and found to be robust across function/problem types~\cite{Bull:1997b}. We used the round-robin approach and partnering with the best individual in our pilot study and return to it here with the focus on learning speed, i.e.,\ how to minimise the number of (timely/costly) fitness evaluations whilst learning effectively.  

As EAs have been applied to ever more complex tasks, surrogate models (also known as meta-models) have been used to reduce the optimisation time. A surrogate model, $y=f(\vec{x})$, can be formed using a sample $\mathcal{D}$ of evaluated designs $N$, where $\vec{x}$ is the genotype describing the design morphology and $y$ is the fitness/performance. The model is then used to compute the fitness of unseen data points $\vec{x} \notin \mathcal{D}$, thereby providing a cheap approximation of the real fitness function for the EA to use. Evaluations with the real fitness function must continue to be performed periodically otherwise the model may lead to premature convergence on local optima (see Jin~\cite{Jin:2011} for an overview). There has been very little prior work on the use of surrogates in a coevolutionary context: they have been shown capable of solving computationally expensive optimisation problems with varying degrees of epistasis more efficiently than conventional coevolutionary EAs (CEAs) through the use of radial basis functions~\cite{Ong:2002} and memetic algorithms~\cite{Goh:2011}. Remarkably, in 1963 Dunham~et~al.~\cite{Dunham:1963}, in describing the evolutionary design of physical logic circuits/devices, briefly note (without giving details): ``It seemed better to run through many `generations' with only approximate scores indicating progress than to manage a very few `evolutions' with rather exact statements of position.'' 

Our aforementioned pilot study is the first known use of coevolutionary design without simulation. As noted above, we have recently compared different modelling techniques by which to construct surrogates for coevolution. In this paper we further consider how best to train and use such models. 

\subsection{The NKCS Model}

Kauffman and Johnsen~\cite{Kauffman:1991} introduced the abstract NKCS model to enable the study of various aspects of coevolution. In their model, an individual is represented by a genome of $N$ (binary) genes, each of which depends epistatically upon $K$ other randomly chosen genes in its genome. Thus increasing $K$, with respect to $N$, increases the epistatic linkage, increasing the ruggedness of the fitness landscapes by increasing the number of fitness peaks, which increases the steepness of the sides of fitness peaks and decreases their typical heights. Each gene is also said to depend upon $C$ randomly chosen traits in each of the other $X$ species with which it interacts, where there are $S$ number of species in total. The adaptive moves by one species may deform the fitness landscape(s) of its partner(s). Altering $C$, with respect to $N$, changes how dramatically adaptive moves by each species deform the landscape(s) of its partner(s). The model assumes all inter- and intragenome interactions are so complex that it is appropriate to assign random values to their effects on fitness. Therefore, for each of the possible $K+(X \times C)$ interactions, a table of $2^{K+(X \times C)+1}$ fitnesses is created for each gene, with all entries in the range $0.0$ to $1.0$, such that there is one fitness for each combination of traits. The fitness contribution of each gene is found from its table; these fitnesses are then summed and normalised by $N$ to give the selective fitness of the total genome for that species. Such tables are created for each species (see example in Figure~\ref{fig:nkcs}; the reader is referred to Kauffman~\cite{Kauffman:1993} for full details). This tuneable model has previously been used to explore coevolutionary optimisation, particularly in the aforementioned comparison of partnering strategies~\cite{Bull:1997b}. We similarly use it here to systematically compare various techniques for the design mining approach.
 
\begin{figure}[t]
	\centering
	\small
	$N=3$~~~~$K=1$~~~~$C=1$~~~~$S=2$~~~~$X=1$\\
	\begin{tikzpicture}
		% first species N
		\node(a) at (0,0)[shape=circle,draw] {$n1$};
		\node(b) at (2,0)[shape=circle,draw] {$n2$};
		\node(c) at (1,-1)[shape=circle,draw] {$n3$};
		\node(astate) at (0,0.6)[] {1};
		\node(bstate) at (2,0.6)[] {0};
		\node(cstate) at (1,-0.4)[] {1};
		\node(label1) at (1,1.5)[] {Species $s1$};
		% second species N
		\node(d) at (5,0)[shape=circle,draw] {$n1$};
		\node(e) at (7,0)[shape=circle,draw] {$n2$};
		\node(f) at (6,-1)[shape=circle,draw] {$n3$};
		\node(dstate) at (5,0.6)[] {1};
		\node(estate) at (7,0.6)[] {1};
		\node(fstate) at (6,-0.4)[] {0};
		\node(label2) at (6,1.5)[] {Species $s2$};
		% first species K
		\draw [black, line width=1pt, ->, >=stealth]
		(a) edge node [above] {} (b)
		(c) edge node [above] {} (a)
		(b) edge node [above] {} (c)
		% second species K
		(d) edge node [above] {} (e)
		(e) edge node [above] {} (d)
		(e) edge node [above] {} (f);
		% first species C
		\draw [black, dashed, line width=1pt, ->, >=stealth]
		(d) edge [bend right=30] node [above] {} (a)
		(f) edge [bend right=5] node [above] {} (b)
		(f) edge [bend right=5] node [above] {} (c)
		% second species C
		(a) edge [bend left=40] node [above] {} (e)
		(b) edge [bend left=5] node [above] {} (d)
		(c) edge [bend right=10] node [above] {} (f);
		% species divider line
%	   	\draw [black, line width=0.5pt] (3.5,1.7) -- (3.5,-1.7);
		% empty space
		\node(em) at (0,-1.6)[] {};
	\end{tikzpicture}
\\
\setlength\tabcolsep{2pt}
	\begin{tabular}{ccc|c}
		\small
		\\
		\multicolumn{4}{c}{Species $s1$ gene $n1$} \\
		\hline
		$s1n1$ & $s1n3$ & $s2n1$ & fitness \\
		\hline
		0 & 0 & 0 & 0.57 \\
		0 & 0 & 1 & 0.12 \\
		0 & 1 & 0 & 0.09 \\
		0 & 1 & 1 & 0.16 \\
		1 & 0 & 0 & 0.44 \\
		1 & 0 & 1 & 0.66 \\
		1 & 1 & 0 & 0.33 \\
		\bf{1} & \bf{1} & \bf{1} & \bf{0.44}\\ [0ex]
	\end{tabular} 
	\hspace{5mm}
	\begin{tabular}{ccc|c}
		\small
		\\
		\multicolumn{4}{c}{Species $s1$ gene $n2$} \\
		\hline
		$s1n2$ & $s1n1$ & $s2n3$ & fitness \\
		\hline
		0 & 0 & 0 & 0.11 \\
		0 & 0 & 1 & 0.32 \\
		\bf{0} & \bf{1} & \bf{0} & \bf{0.68} \\
		0 & 1 & 1 & 0.30 \\
		1 & 0 & 0 & 0.19 \\
		1 & 0 & 1 & 0.77 \\
		1 & 1 & 0 & 0.21 \\
		1 & 1 & 1 & 0.23 \\ [0ex]
	\end{tabular}
	\hspace{5mm}
	\begin{tabular}{ccc|c}
		\small
		\\
		\multicolumn{4}{c}{Species $s1$ gene $n3$} \\
		\hline
		$s1n3$ & $s1n2$ & $s2n3$ & fitness \\
		\hline
		0 & 0 & 0 & 0.75\\
		0 & 0 & 1 & 0.42\\
		0 & 1 & 0 & 0.25\\
		0 & 1 & 1 & 0.28\\
		\bf{1} & \bf{0} & \bf{0} & \bf{0.13}\\
		1 & 0 & 1 & 0.58\\
		1 & 1 & 0 & 0.66\\
		1 & 1 & 1 & 0.91\\ [0ex]
	\end{tabular}
	\caption{The NKCS model: Each gene is connected to $K$ randomly chosen local genes (solid lines) and to $C$ randomly chosen genes in each of the $X$ other species (dashed lines). A random fitness is assigned to each possible set of combinations of genes. The fitness of each gene is summed and normalised by $N$ to give the fitness of the genome. An example NKCS model is shown above and example fitness tables are provided for species $s1$, where the $s1$ genome fitness is 0.416 when $s1=[101]$ and $s2=[110]$.}
	\label{fig:nkcs}
\end{figure}
 
That is, each species is cast as a sub-thread of an overall design task, thereby enabling examination of the effects from varying their number ($S$), their individual complexity ($K$), and the degree of interdependence between them ($C$). The fitness calculations of each species are combined to give a global system performance.

\subsection{Evolving Wind Farms}  

As we have reviewed elsewhere~\cite{Preen:2016}, techniques such as EAs have been used to design wind turbine blades using CFD simulations, some in conjunction with surrogate models, e.g., Chen~et~al.~\cite{Chen:2012}. EAs have also been extensively used to optimise the turbine positioning within wind farms, e.g., Mosetti~et~al.~\cite{Mosetti:1994}. Most work has focused on arrays of homogeneous turbines, however wind farms of heterogeneous height have recently gained attention as a means to improve the overall power output for a given number of turbines~\cite{Chen:2013,DuPont:2016,Chen:2016}. Chamorro~et~al.~\cite{Chamorro:2014} explored horizontal-axis wind turbine (HAWT) farms with large and small turbines positioned alternately. They found that size heterogeneity has positive effects on turbulent loading as a result of the larger turbines facing a more uniform turbulence distribution and the smaller turbines operating under lower turbulence levels. Craig~et~al.~\cite{Craig:2016} have demonstrated a similar potential for heterogeneous height VAWT wind farms. Chowdhury~et~al.~\cite{Chowdhury:2012} optimised layouts of HAWT with heterogeneous rotor diameters using particle swarm optimisation and found that the optimal combination of turbines with differing rotor diameters significantly improved the wind farm efficiency. Recently, Xie~et~al.~\cite{Xie:2016} have performed simulations of wind farms with collocated VAWT and HAWT, showing the potential to increase the efficiency of existing HAWT wind farms by adding VAWTs.

Conventional offshore wind farms require support structures fixed rigidly to the seabed, which currently limits their deployment to depths below 50 m. However, floating wind farms can be deployed in deep seas where the wind resources are strongest, away from shipping lanes and wind obstructions~\cite{Paulsen:2012}. See Borg~et~al.~\cite{Borg:2014} for a recent review of floating wind farms. They note that floating VAWT have many advantages over HAWT, e.g., lower centre of gravity, increased stability, and increased tolerance of extreme conditions. The design of floating wind farms is especially challenging since platform oscillations also need to be considered. EAs are beginning to be used to explore the design of floating support structures, e.g., Hall~et~al.~\cite{Hall:2013} optimised HAWT platforms using a simple computational model to provide fitness scores. Significantly, all of these works have involved the use of CFD simulations with varying degrees of fidelity.

Our pilot study found that asymmetrical pairs of VAWTs can be more efficient than similar symmetrical designs. In this paper, we extend our initial work to the heterogeneous design of an array of 6 closely positioned VAWT, which is currently effectively beyond the capabilities of accurate 3D CFD simulation; the approach performs optimisation in the presence of non-uniform wind velocity, complex inter-turbine wake effects, and multi-directional wind flow from nearby obstacles, which is extremely difficult to achieve accurately under high fidelity CFD simulation. In addition, previously the combined rotational speed was simply used as the objective measure, whereas here we use the total angular kinetic energy of the array, which includes both mass and speed of rotation, and we use a more flexible spline representation that enables the potential exploitation of both drag and lift forces in conjunction with inter-turbine flow and turbulence from nearby obstacles. 

\section{Surrogate-assisted Coevolution}

The basic coevolutionary genetic algorithm (CGA) is outlined in Algorithm~\ref{alg:cga}. Initially all individuals in each of the species/populations must be evaluated. Since no initial fitness values are known, a random individual is chosen in each of the other populations to form a global solution, however if there is a known good individual then that individual can be used instead. The CGA subsequently cycles between populations, selecting parents via tournament and creating offspring with mutation and/or crossover. The offspring are then evaluated using representative members from each of the other populations. At any point during evolution, each individual is assigned the maximum team fitness achieved by any team in which it has been evaluated, where the team fitness is the sum of the fitness scores of each collaborating member.

For the basic surrogate-assisted CGA (SCGA) used in this paper, the CGA runs as normal except that each time a parent is chosen, $\lambda_{m}$ number of offspring are created and then evaluated with an artificial neural network surrogate model; the single offspring with the highest approximated fitness is then evaluated on the real fitness function in collaboration with the fittest solution (best partner) in each other populations. See outline in Algorithm~\ref{alg:scga}. The model is trained using backpropagation for $T$ epochs; where an epoch consists of randomly selecting, without replacement, all individuals from a species population archive and updating the model weights at a learning rate $\beta$. The model weights are (randomly) reinitialised each time before training due to the temporal nature of the collaborating scheme.  
 
\begin{algorithm}[t]
	\small
	\SetAlgoLined%
	\For{each species}{
		initialise population\;
		select a random representative for each other species\;
		\For{each individual in population}{
			evaluate\;
		}
	}
	\While{evaluation budget not exhausted}{
		\For{each species}{
			create an offspring using genetic operators\;
			select a representative for each other species\;
			evaluate the offspring\;
			add offspring to species population\;
		}
	}
	\caption{Coevolutionary genetic algorithm}
	\label{alg:cga}
\end{algorithm}

\begin{algorithm}[t]
	\small
	\SetAlgoLined%
	\For{each species}{
		initialise population\;
		select a random representative for each other species\;
		\For{each individual in population}{
			evaluate\;
			archive\;
		}
	}
	\While{evaluation budget not exhausted}{
		\For{each species}{
			initialise model\;
			train model on species archive\;
			select parent(s) using tournament selection\;
			\For{$\lambda_{m}$ number of times}{
				create an offspring using genetic operators\;
				predict offspring fitness using the model\;
			}
			select the offspring with the highest model predicted fitness\;
			select a representative for each other species\;
			evaluate the offspring\;
			add offspring to species population\;
			archive offspring\;
		}
	}
	\caption{Surrogate-assisted coevolutionary genetic algorithm}
	\label{alg:scga}
\end{algorithm}
 
For both CGA and SCGA, a tournament size of 3 takes place for both selection and replacement. A limited form of elitism is used whereby the current fittest member of the population is given immunity from deletion.   

\section{NKCS Experimentation}

\begin{figure}[h]
	\centering 
	\small
	\begin{tikzpicture}
		\node(a) at (0.0,0)[shape=circle,draw] {$s1$};
		\node(b) at (1.5,0)[shape=circle,draw] {$s2$};
		\node(c) at (3.0,0)[shape=circle,draw] {$s3$};
		\node(d) at (4.5,0)[shape=circle,draw] {$s4$};
		\node(e) at (6.0,0)[shape=circle,draw] {$s5$};
		\node(f) at (7.5,0)[shape=circle,draw] {$s6$};
		\draw [black, line width=1pt, <->, >=stealth]
		(a) edge node [above] {} (b)
		(b) edge node [above] {} (c)
		(c) edge node [above] {} (d)
		(d) edge node [above] {} (e)
		(e) edge node [above] {} (f);
	\end{tikzpicture}
	\caption{NKCS topology. Arrows indicate inter-species connectivity ($X$).}
	\label{fig:nkcs_topology}
\end{figure}
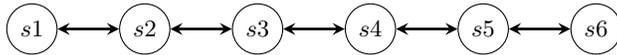

For the physical experiments performed in this paper, 6 VAWT are positioned in a row. Therefore, to simulate this interacting system, we explore the case where $S=6$ and each species is affected by its proximate neighbours, i.e., $X=1$ for the first and sixth species, and $X=2$ for all others. Figure~\ref{fig:nkcs_topology} illustrates the simulated topology. For all NKCS simulations performed, $P=20$, $N=20$, per allele mutation probability $\mu=5\%$, and crossover probability is $0\%$. Where a surrogate model is used, the model parameters are: $N$ input neurons, $H=10$ hidden neurons, 1 output neuron, $\lambda_{m}=1000$, $T=50$, $\beta=0.1$. All results presented are an average of 100 experiments consisting of 10 coevolutionary runs on 10 randomly generated NKCS functions. The performance of all algorithms are shown on four different $K$ and $C$ values, each representing a different point in the range of inter- and intra-population dependence. 
 
\subsection{Coevolution}
 
We begin by comparing the traditional approach of partnering with the elite member in each other species (CGA-b) with performing additional evaluations (CGA-br), and explore any benefits to overall learning speed from refreshing the population fitness values as the fitness landscapes potentially shift; that is, all individuals in the other species populations are re-evaluated in collaboration with the current elite members each time a new fittest individual is found (CGA-re). 

As noted above, after Potter and De~Jong~\cite{Potter:1994}, traditionally CEAs consider each population in turn. Thus, if $S=10$ and each species population creates one offspring per turn, then 10 evaluations are required for the whole system. However, at the other end of this scale each population simultaneously generates a new individual each turn, and evaluates all offspring at once~\cite{Bull:1993}, therefore requiring only one evaluation for the whole system. Varying the number of offspring collaborators in this way is much like varying the system-level mutation rate. Therefore, we explore the case where all species offspring are created and tested simultaneously (CGA-o).

In summary, the NKCS model is used to examine the following four different collaboration schemes:

\begin{itemize}
  \item CGA-b: each offspring is evaluated in collaboration with the current best individual in each of the other species populations.
  \item CGA-br: each offspring is evaluated as in CGA-b, and additionally with a random member in each of the other populations.
  \item CGA-re: each offspring is evaluated as in CGA-b, and all populations are re-evaluated when one makes a progress.
  \item CGA-o: offspring are created in each species simultaneously and evaluated together.
\end{itemize}

\begin{figure*}[t]
	\centering 
	\includegraphics[width=\linewidth]{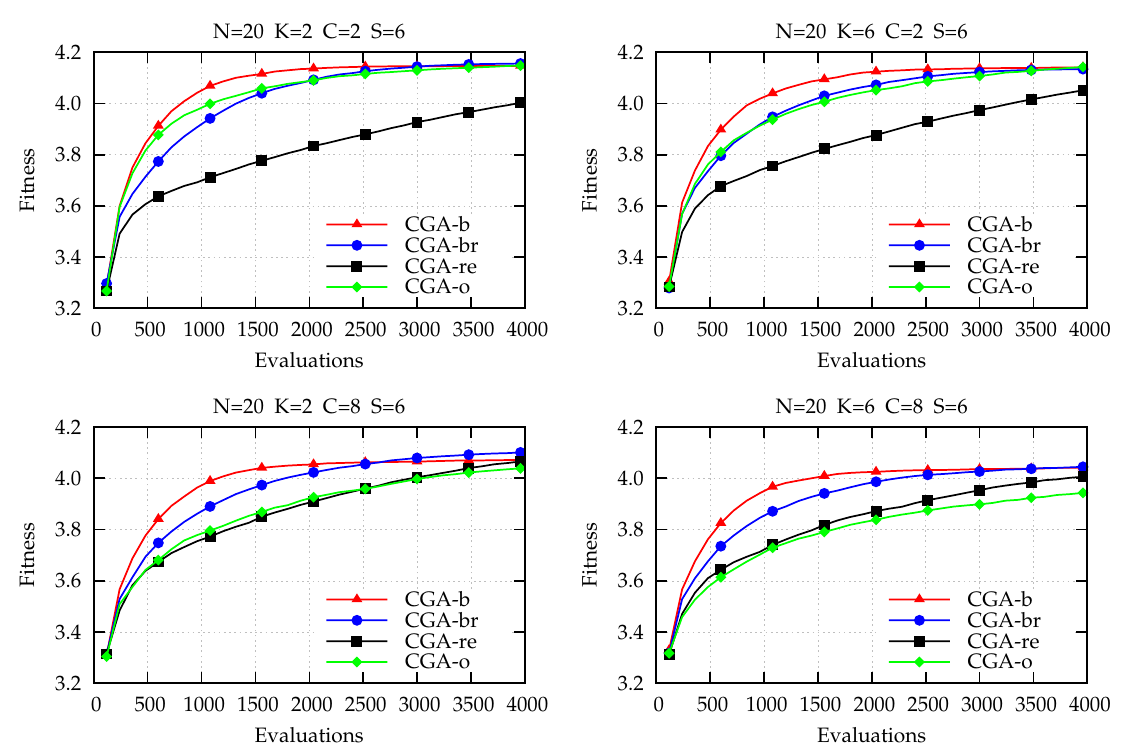}
	\caption{CGA mean best fitness. Results are an average of 100 experiments consisting of 10 coevolutionary runs of 10 random NKCS functions. CGA-b (triangle), CGA-br (circle), CGA-re (square), and CGA-o (diamond).}
	\label{fig:cga}
\end{figure*}

Figure~\ref{fig:cga} and Table~\ref{table:cga_stats_s6} present the performance of the collaboration schemes. As can be seen, during the early stages of evolution, the mean best fitness of CGA-b is significantly greater than CGA-br and CGA-re for all tested $K,C$ values, showing that performing additional evaluations results in a lower fitness compared with the approach of only collaborating with the elite members. At the end of the experiments, the three approaches generally reach approximately similar performance, suggesting that there is no penalty for this increase in early learning speed. For the case of both lower inter- and intra-population epistasis CGA-br performs better, which supports findings reported elsewhere~\cite{Potter:1994,Bull:1997b}. The approach of evaluating all offspring simultaneously (CGA-o) appears to be detrimental to performance under the simulated conditions.
  
\begin{table}[t]
    \centering
	\captionsetup{justification=centering}
	\caption{CGA best fitnesses after 480 and 3600 evaluations (averages of 100). The mean is highlighted in boldface where it is significantly different from CGA-b using a Mann-Whitney $U$ test at the 95\% confidence interval.}
    \begin{tabular}{l r r r r }
		\toprule
		& CGA-b & CGA-br & CGA-re & CGA-o \\
		\hline
		\multicolumn{5}{l}{After 480 evaluations:} \\
		$K2C2$ & 3.8449 & {\bf 3.7141} & {\bf 3.6066} & {\bf 3.8163} \\
		$K2C8$ & 3.7767 & {\bf 3.6937} & {\bf 3.6387} & {\bf 3.6424} \\
		$K6C2$ & 3.8338 & {\bf 3.7359} & {\bf 3.6423} & {\bf 3.7617} \\
		$K6C8$ & 3.7626 & {\bf 3.6765} & {\bf 3.6105} & {\bf 3.5761} \\
		\multicolumn{5}{l}{After 3600 evaluations:} \\
		$K2C2$ & 4.1464 & 4.1536 & {\bf 3.9757} & 4.1417 \\
		$K2C8$ & 4.0700 & 4.0949 & {\bf 4.0469} & {\bf 4.0269} \\
		$K6C2$ & 4.1395 & 4.1321 & {\bf 4.0254} & 4.1320 \\
		$K6C8$ & 4.0390 & 4.0403 & {\bf 3.9926} & {\bf 3.9279} \\
		\bottomrule
    \end{tabular}
    \label{table:cga_stats_s6}
\end{table}

\subsection{Surrogate-assisted Coevolution}

In this section we compare the performance of CGA-b with the standard surrogate-assisted version (SCGA-b). In addition, we compare the performance of the standard surrogate approach where the models are presented only the $N$ genes from their own species (SCGA-b) with the case where the models are presented all $N\times S$ partner genes (SCGA-a). We also compare the standard approach of evaluating the most promising of $\lambda_{m}$ offspring stemming from a single parent (SCGA-b) with searching the same number of offspring where $\lambda_{m}$ tournaments are performed to select parents that each create a single offspring (SCGA-p).

Furthermore, due to the highly temporal nature of the individuals undergoing evaluation needing to partner with the elite members in each of the other species, it is possible that the surrogate model performance may degrade by using the entire data set for training. For example, individuals from the initial population may perform very differently when partnered with the elite individuals from later generations. A windowed approach of using only the most recent $P$ evaluated individuals in each species for training seemed promising in a prior experiment coevolving a pair of VAWT, however was not statistically significant in practice~\cite{Preen:2016}. Here we explore the effect for larger numbers of species where the temporal variance is potentially much higher (SCGA-bw).

In summary, the algorithms tested:
 
\begin{itemize}
  \item SCGA-b: standard SCGA.
  \item SCGA-a: global surrogate model construction.
  \item SCGA-p: $\lambda_{m}$ parents are selected via tournaments, each creating a single offspring and the most promising as suggested by the model is evaluated.
  \item SCGA-bw: most recent $P$ evaluated individuals used for training.
\end{itemize}        
                                                        
\begin{figure*}[t]
	\centering 
	\includegraphics[width=\linewidth]{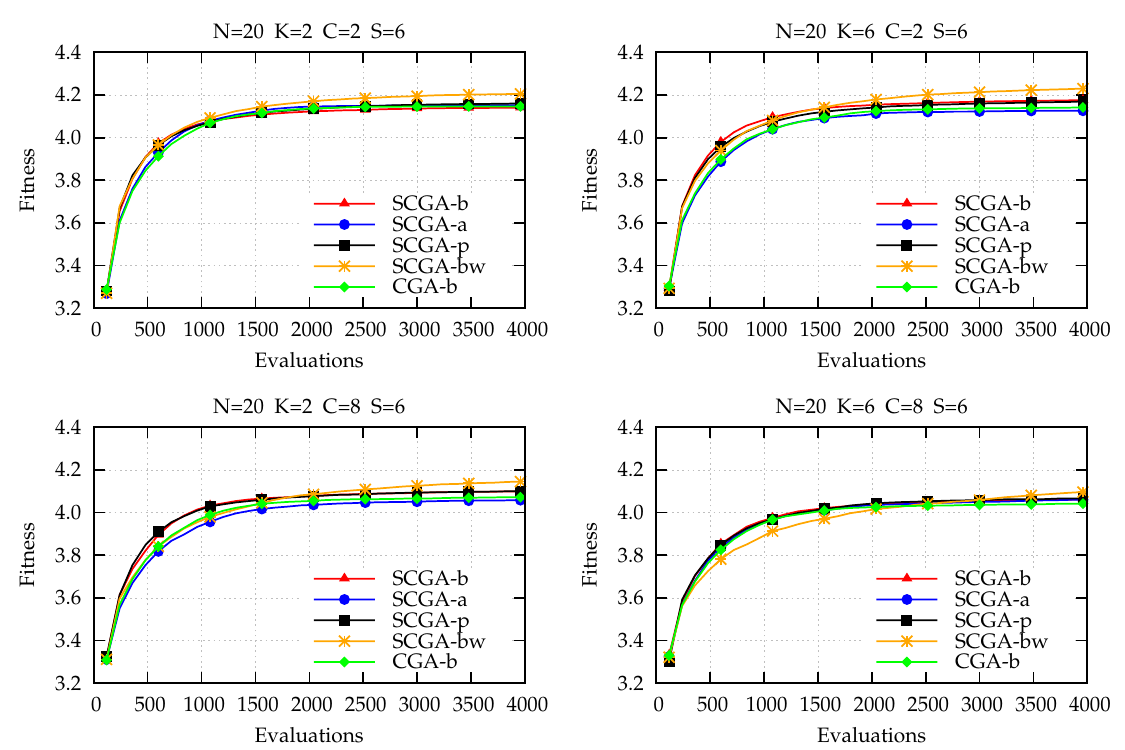}
	\caption{SCGA mean best fitness. Results are an average of 100 experiments consisting of 10 coevolutionary runs of 10 random NKCS functions. SCGA-b (triangle), SCGA-a (circle), SCGA-p (square), SCGA-bw (star), and CGA-b (diamond).}
	\label{fig:scga}
\end{figure*}

The results are presented in Figure~\ref{fig:scga} and Table~\ref{table:scga_stats_s6}. As can be seen, the use of the surrogate model to identify more promising offspring clearly increases learning early in the search. For example, the mean best fitness of SCGA-b is significantly greater for all tested $K,C$ values with the exception of very high inter- and intra-population epistasis. At the end of the experiments, similar optima are reached, showing that there is no penalty for this increase in early learning speed. The benefit of the divide-and-conquer strategy to model building can be seen by comparing SCGA-b with SCGA-a. The mean best fitness of SCGA-b is significantly greater than SCGA-a for all four $K,C$ values after 480 evaluations, with the exception of very high $K$ and $C$; showing that purely local models are both efficient and scalable. Comparing SCGA-b with SCGA-p shows that the simple method of using the model we presented in our pilot study is quite robust as there is no significant difference. Finally, the windowed training scheme (SCGA-bw) was found to be significantly worse than using all data (SCGA-b) during the early stages of evolution, however later in the experiments reached a higher optima.
 
\begin{table}[t]
    \centering
	\captionsetup{justification=centering}
	\caption{SCGA best fitnesses after 480 and 3600 evaluations (averages of 100). The mean is highlighted in boldface where it is significantly different from SCGA-b using a Mann-Whitney $U$ test at the 95\% confidence interval.}
    \begin{tabular}{l r r r r r}
		\toprule
		& SCGA-b & CGA-b & SCGA-a & SCGA-p & SCGA-bw \\
		\hline
		\multicolumn{5}{l}{After 480 evaluations:} \\
		$K2C2$ & 3.9094 & {\bf 3.8449} & {\bf 3.8633} & 3.9070 & 3.9071 \\
		$K2C8$ & 3.8214 & {\bf 3.7767} & {\bf 3.7537} & 3.8477 & {\bf 3.7778} \\
		$K6C2$ & 3.9160 & {\bf 3.8338} & {\bf 3.8175} & 3.8987 & {\bf 3.8794} \\
		$K6C8$ & 3.7847 & 3.7626 & 3.7750 & 3.7851 & {\bf 3.7252} \\
		\multicolumn{5}{l}{After 3600 evaluations:} \\
		$K2C2$ & 4.1392 & 4.1464 & 4.1521 & 4.1578 & {\bf 4.2027} \\
		$K2C8$ & 4.0974 & 4.0700 & {\bf 4.0558} & 4.0970 & {\bf 4.1383} \\
		$K6C2$ & 4.1733 & {\bf 4.1395} & {\bf 4.1254} & 4.1657 & {\bf 4.2244} \\
		$K6C8$ & 4.0596 & 4.0390 & 4.0571 & 4.0630 & {\bf 4.0849} \\
		\bottomrule
    \end{tabular}
	\label{table:scga_stats_s6}
\end{table}
 
\section{VAWT Wind Farm Design}

\subsection{Methodology}

A single 2-stage 2-blade VAWT candidate with end plates is here created as follows. End plates are drawn in the centre of a Cartesian grid with a diameter of 35~mm and thickness of 1~mm. A central shaft 70~mm tall, 1~mm thick, and with a 1~mm hollow diameter is also drawn in the centre of the grid in order to mount the VAWT for testing. The 2D profile of a single blade on 1 stage is represented using 5 ($x,y$) co-ordinates on the Cartesian grid, i.e., 10 genes, $x_1,y_1, ..., x_5,y_5$. A spline is drawn from ($x_1,y_1$) to ($x_3,y_3$) as a quadratic B\'{e}zier curve, with ($x_2,y_2$) acting as the control point. The process is then repeated from ($x_3,y_3$) to ($x_5,y_5$) using ($x_4,y_4$) as control. The thickness of the spline is a fixed size of 1~mm. The co-ordinates of the 2D blade profile are only restricted by the plate diameter; that is, the start and end position of the spline can be located anywhere on the plate. 

To enable $z$-axis variation, 3 additional co-ordinates (i.e., 6 genes) are used to compute cubic B\'{e}zier curves in the $xz$ and $yz$ planes that offset the 2D profile. The $xz$-plane offset curve is formed from an $x$ offset=0 on the bottom plate to an $x$ offset=0 on the top plate using control points ($zx_1,z_1$) and ($zx_2,z_2$). The $yz$-plane offset curve is formed in the same way with $zy_1$ and $zy_2$ control points, however reusing $z_1$ and $z_2$ to reduce the number of parameters to optimise. 

\begin{figure}[t]
	\centering 
	\subfigure[Seed design; genome: $x_1=15.1$, $y_1=15.1$, $x_2=22.1$, $y_2=15.1$, $x_3=25.7$, $y_3=15.9$, $x_4=32.1$, $y_4=16.1$, $x_5=32.1$, $y_5=27.1$, $zx_1=0$, $zx_2=0$, $zy_1=0$, $zy_2=0$, $z_1=20$, $z_2=27.2$, $r_1=0$.]{%
		\includegraphics[width=0.22\linewidth]{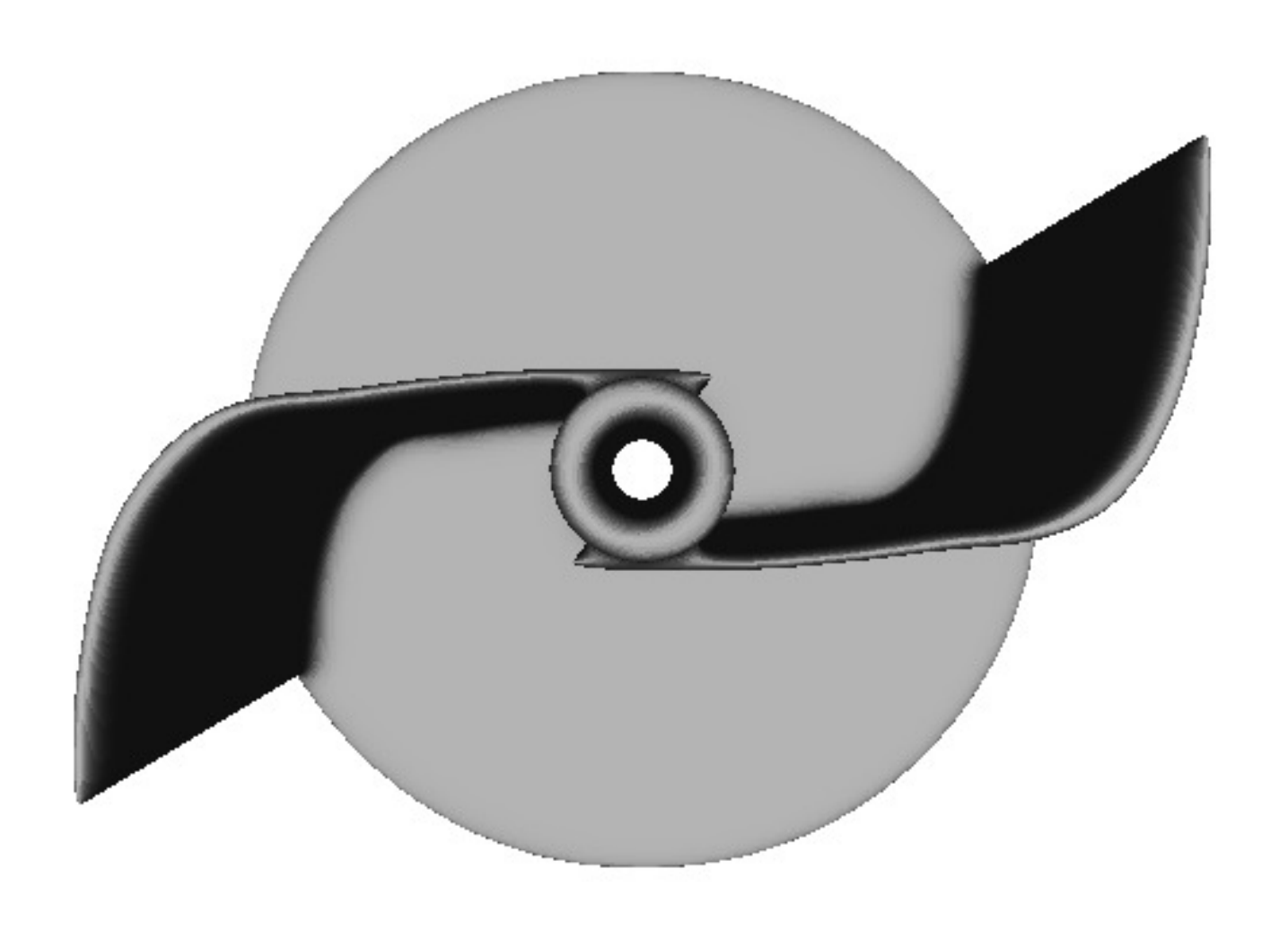}%
	\includegraphics[width=0.22\linewidth]{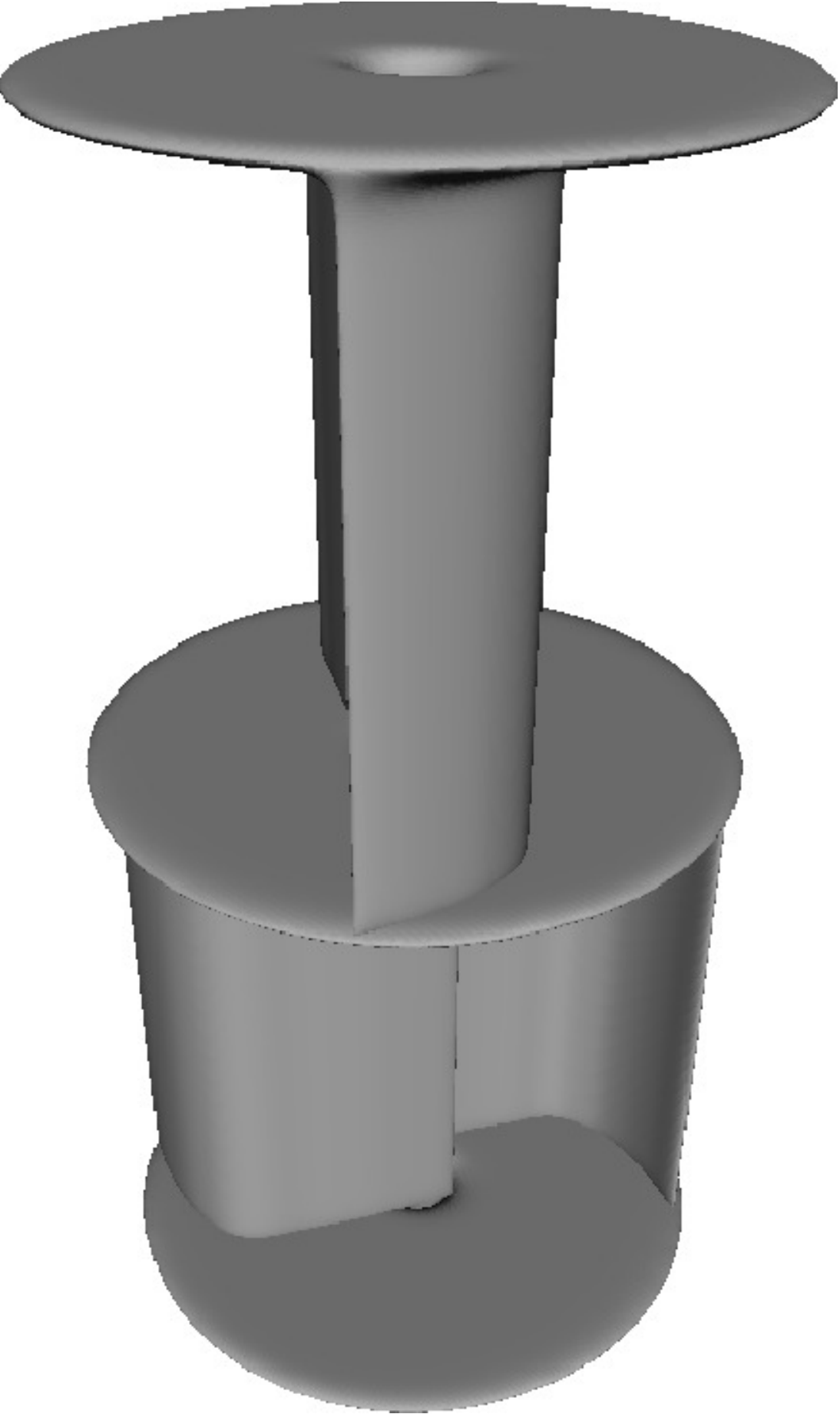}\label{fig:seed-cad} } \hspace{5mm}
	\subfigure[Seed design printed by a 3D printer; 35~mm diameter; 70~mm tall; 7~g; 45-min printing time at 0.3~mm resolution.] {%
		\includegraphics[width=0.22\linewidth]{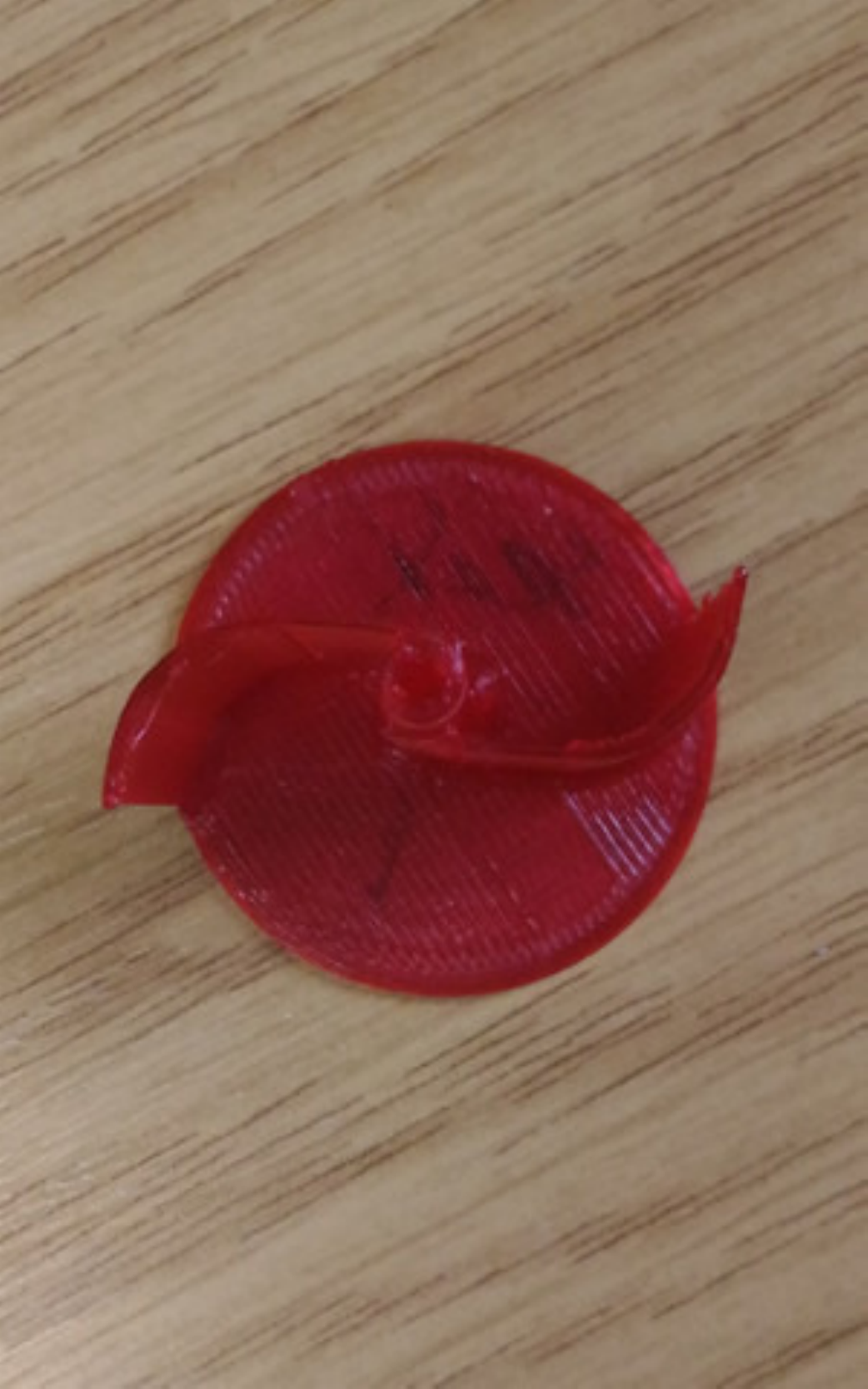}%
	\includegraphics[width=0.22\linewidth]{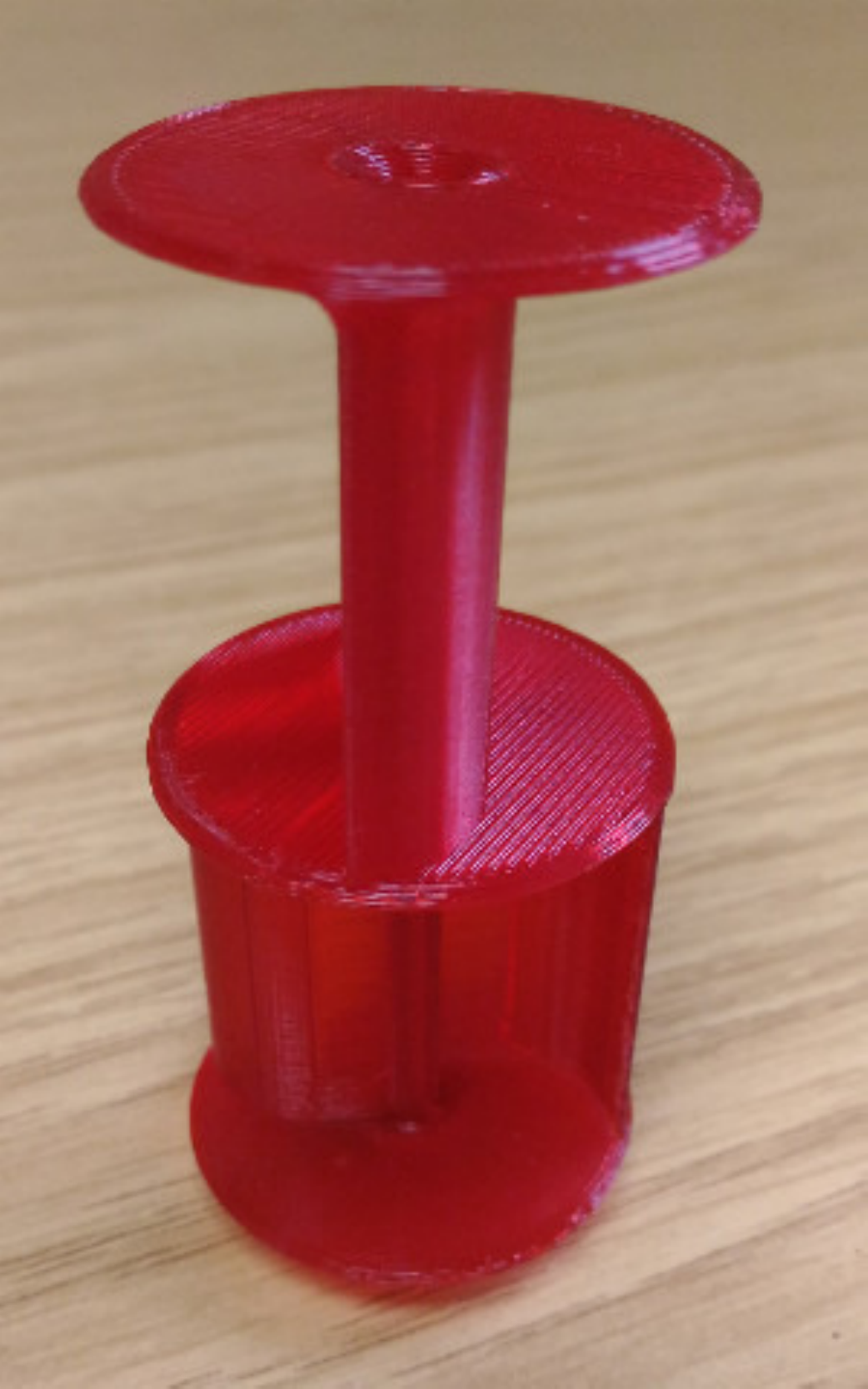}\label{fig:seed-printed} }
	\caption{Example VAWT.}
\end{figure}

Furthermore, an extra gene, $r_1$, specifies the degree of rotation whereby the blades of 1-stage are rotated from one end plate to the next through the $z$-axis to ultimately $0-180\degree$. Thus, a total of 17 genes specify the design of a candidate VAWT. The blade is then duplicated and rotated $180\degree$ to form a 2-bladed design. The entire stage is then duplicated and rotated $90\degree$ to form the second stage of the VAWT; see example design in Figure~\ref{fig:seed-cad}. When physical instantiation is required, the design is fabricated by a 3D printer (Replicator~2; MakerBot Industries LLC, USA) using a polylactic acid (PLA) bioplastic at 0.3~mm resolution. Figure~\ref{fig:seed-printed} shows the VAWT after fabrication.

In order to provide sufficient training data for the surrogate model, initially CGA-b proceeds for 3 generations before the model is used, i.e., a total of 360 physical array evaluations with 60 evaluated individuals in each species. $S=6$ species are explored, each with $P=20$ individuals, a per allele mutation probability, $\mu=25\%$ with a mutation step size, $\sigma_1=3.6$~(mm) for co-ordinates and $\sigma_2=18\degree$ for $r_1$, and a crossover probability of 0\%. Each species population is initialised with the example design in Figure~\ref{fig:seed-cad} and 19 variants mutated with $\mu=100\%$. The individuals in each species population are initially evaluated in collaboration with the seed individuals in the other species populations. Thereafter, CGA-b runs as normal by alternating between species after a single offspring is formed and evaluated with the elite members from the other species. After 3 generations, SCGA-b is used for an additional generation. To explore whether there is any benefit in windowing the training data, the SCGA is subsequently rerun for 1 generation starting with the same previous CGA-b populations, however using only the current species population for model training (SCGA-bw). The model parameters: 17 input neurons, $H=10$ hidden neurons, 1 output neuron, $\lambda_{m}=1000$, $T=1000$, $\beta=0.1$. Each VAWT is treated separately by evolution and approximation techniques, i.e.,\ heterogeneous designs could therefore emerge.
 
The fitness, $f$, of each individual is the total angular kinetic energy of the collaborating array, 
\begin{equation}
f = \sum\limits_{i=1}^S KE_i
\end{equation}
where the angular kinetic energy (J), $KE$, of an individual VAWT,
\begin{equation}
KE = \frac{1}{2}I\omega^2
\end{equation}
with angular velocity (rad/s), $\omega=\frac{rpm}{60}2\pi$, and moment of inertia (kg$\cdot$m$^2$), $I=\frac{1}{2}mr^2$ with $m$ mass (kg), and $r$ radius (m). 
                                                                                           
The rotational speed (rpm) is here measured using a digital photo laser tachometer (PCE-DT62; PCE Instruments UK Ltd) by placing a $10\times2$~mm strip of reflecting tape on the outer tip of each VAWT and recording the maximum achieved over the period of $\sim30$~s during the application of wind generated by a propeller fan. 

\begin{figure}[t]
	\centering 
	\includegraphics[width=0.8\linewidth]{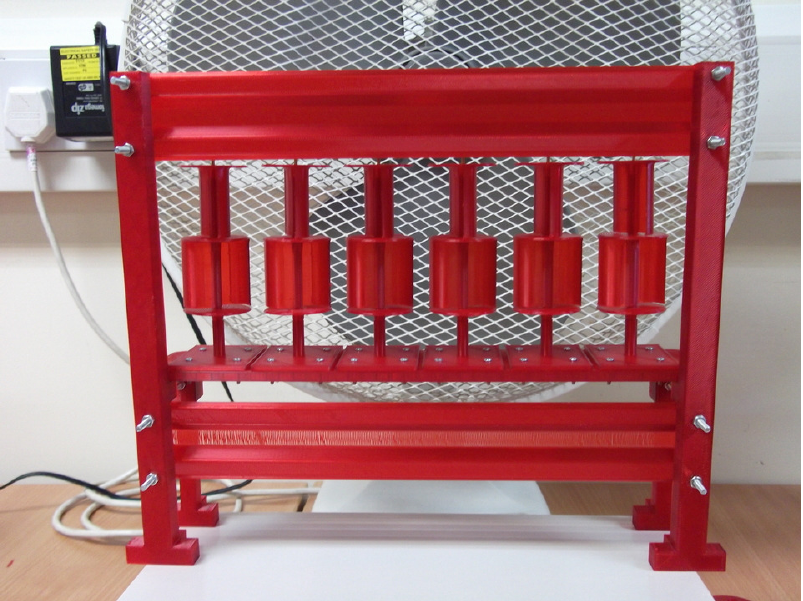}
	\caption{Experimental setup with 6 VAWT. Frame width 275~mm, height 235~mm. Vertical support pillars $10\times15\times235$~mm. Upper and lower cross bars each with height 40~mm, thickness 1~mm, and protruding 8~mm. VAWT freely rotate on rigid metal pins 1~mm in diameter and positioned 42.5~mm adjacently.}
	\label{fig:setup-fan}
\end{figure}

Figure~\ref{fig:setup-fan} shows the test environment with the 30~W, 3500~rpm, 304.8~mm propeller fan, which generates 4.4~m/s wind velocity, and 6 turbines mounted on rigid metal pins 1~mm in diameter and positioned 42.5~mm adjacently and 30~mm from the propeller fan. That is, there is an end plate separation distance of 0.2 diameters between turbines. It is important to note that the wind generated by the fan is highly turbulent with non-uniform velocity and direction across the test platform, i.e., each turbine position receives a different amount of wind energy from different predominant directions and wind reflecting from the test frame may cause multi-directional wind flow. Thus, the designs evolved under such conditions will adapt to these exact environmental circumstances.
 
\subsection{Results}

Each generation consisted of 120 fabrications and array evaluations. After evaluating all individuals in the initial species populations, no mutants were found to produce a greater total kinetic energy than the seed array. After 1 evolved CGA-b generation, the fittest array combination generated a greater total kinetic energy of 7.6~mJ compared with the initial seed array, which produced 5.9~mJ. A small increase in total mass of 42~g to 44.8~g was also observed. After 2 evolved CGA-b generations, the fittest array generated a total kinetic energy of 10~mJ with a further small increase in total mass to 45.6~g. SCGA-b was then used for 1 additional generation and produced a total kinetic energy of 12.2~mJ with a further increase in mass to 49.3~g. 

The fittest SCGA-bw array produced a greater total kinetic energy of 14.8~mJ than SCGA-b. Furthermore, the SCGA-bw mean kinetic energy ($M=12.27$, $SD=1.29$, $N=120$) was significantly greater than SCGA-b ($M=10.56$, $SD=0.93$, $N=120$) using a two-tailed Mann-Whitney test ($U=2076$, $p\le0.001$), showing that windowing the model training data was found to be beneficial in this experiment. SCGA-b and SCGA-bw appear to have predominantly exploited different components of the fitness measure, with SCGA-b finding heavier turbine designs (+8\%) that maintain approximately the same rpm (+3\%), whereas SCGA-bw discovered designs that were approximately the same mass (+0.6\%) with significantly increased rpm (+22\%). 

\begin{figure}[t]
	\centering 
	\subfigure[Array angular kinetic energy.] {
		\includegraphics[width=0.48\linewidth]{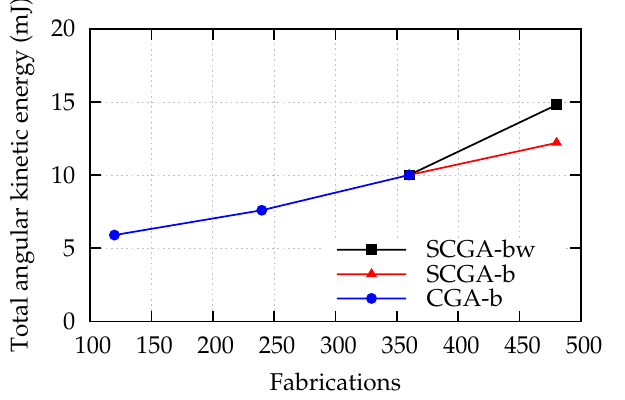}
		\label{fig:ke}
	}
	\subfigure[Array mass.] {
		\includegraphics[width=0.48\linewidth]{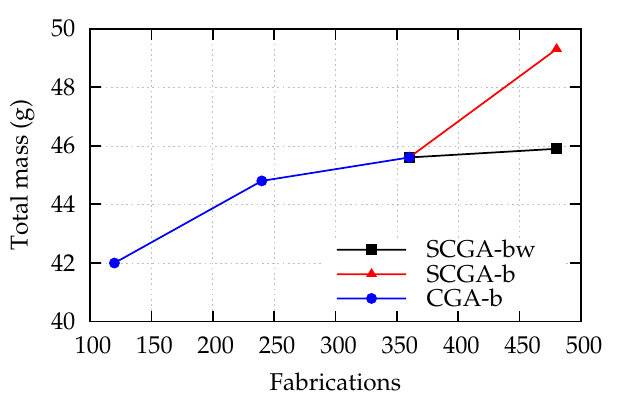}
		\label{fig:mass}
	} \\
	\subfigure[Array rotation speed.] {
		\includegraphics[width=0.48\linewidth]{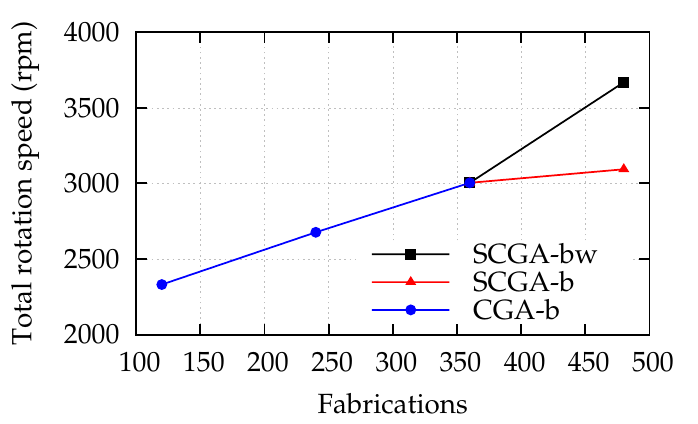}
		\label{fig:rpm}
	}
	\caption{Performance of the fittest evolved VAWT arrays. CGA-b (circle), SCGA-b (triangle), SCGA-bw (square). The SCGAs are used only after 360 evaluations (i.e., 3 generations) of the CGA since sufficient training data is required.}
\end{figure}

\begin{figure*}[t]
	\centering 
	\subfigure[s1]{\includegraphics[width=2.0cm]{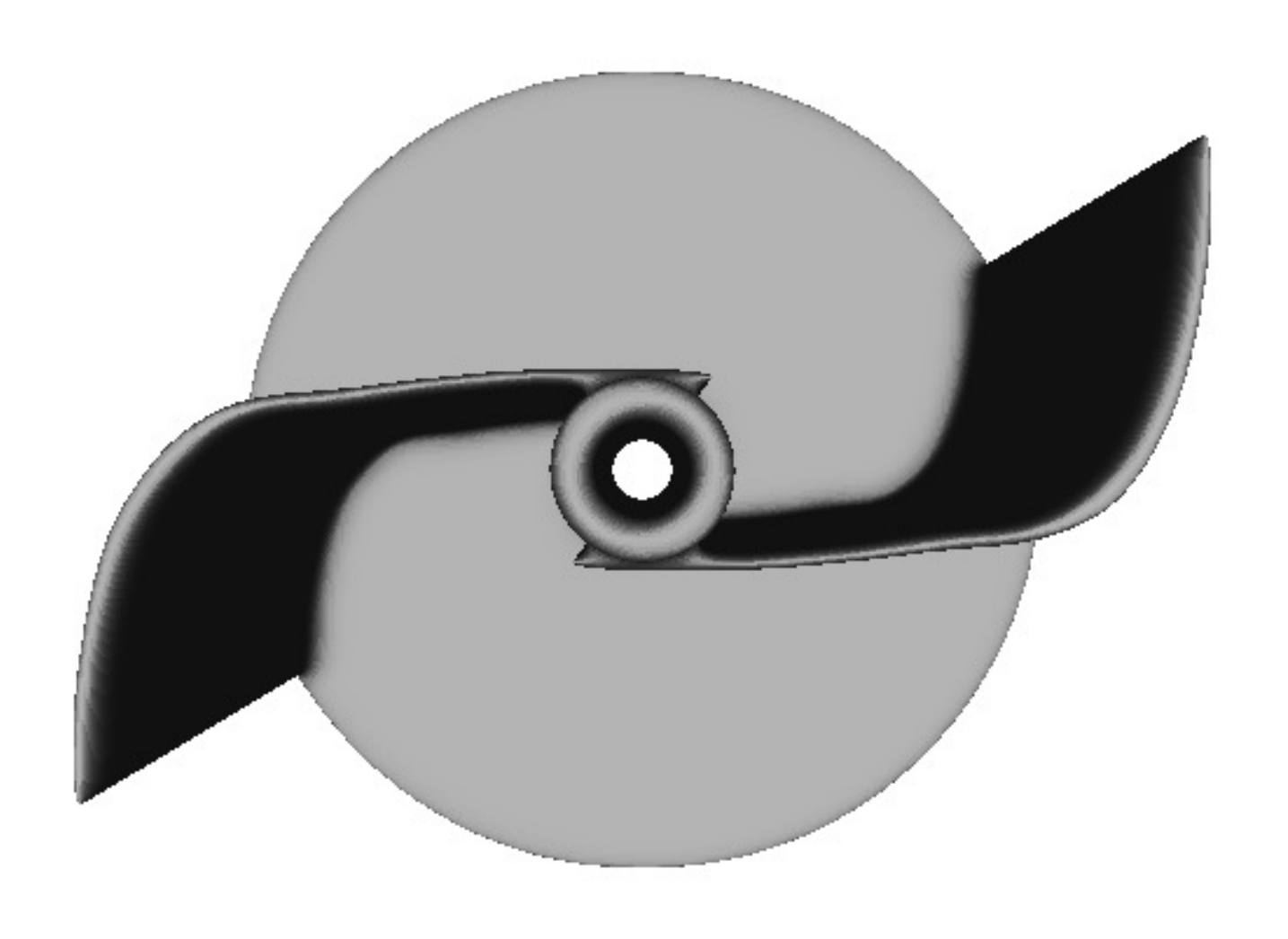}}
	\subfigure[s2]{\includegraphics[width=2.0cm]{fig8.pdf}}
	\subfigure[s3]{\includegraphics[width=2.0cm]{fig8.pdf}}
	\subfigure[s4]{\includegraphics[width=2.0cm]{fig8.pdf}}
	\subfigure[s5]{\includegraphics[width=2.0cm]{fig8.pdf}}
	\subfigure[s6]{\includegraphics[width=2.0cm]{fig8.pdf}}
	\caption{Cross sections of the fittest VAWT array after 1 CGA-b generation, i.e., initial population. No mutants resulted in greater total $KE$ than the seed array. Total $KE=5.9$~mJ, $m=42$~g, 2332~rpm.}
	\label{fig:cga-1gen}
\end{figure*}

\begin{figure*}[t]
	\centering 
	\subfigure[s1]{\includegraphics[width=2.0cm]{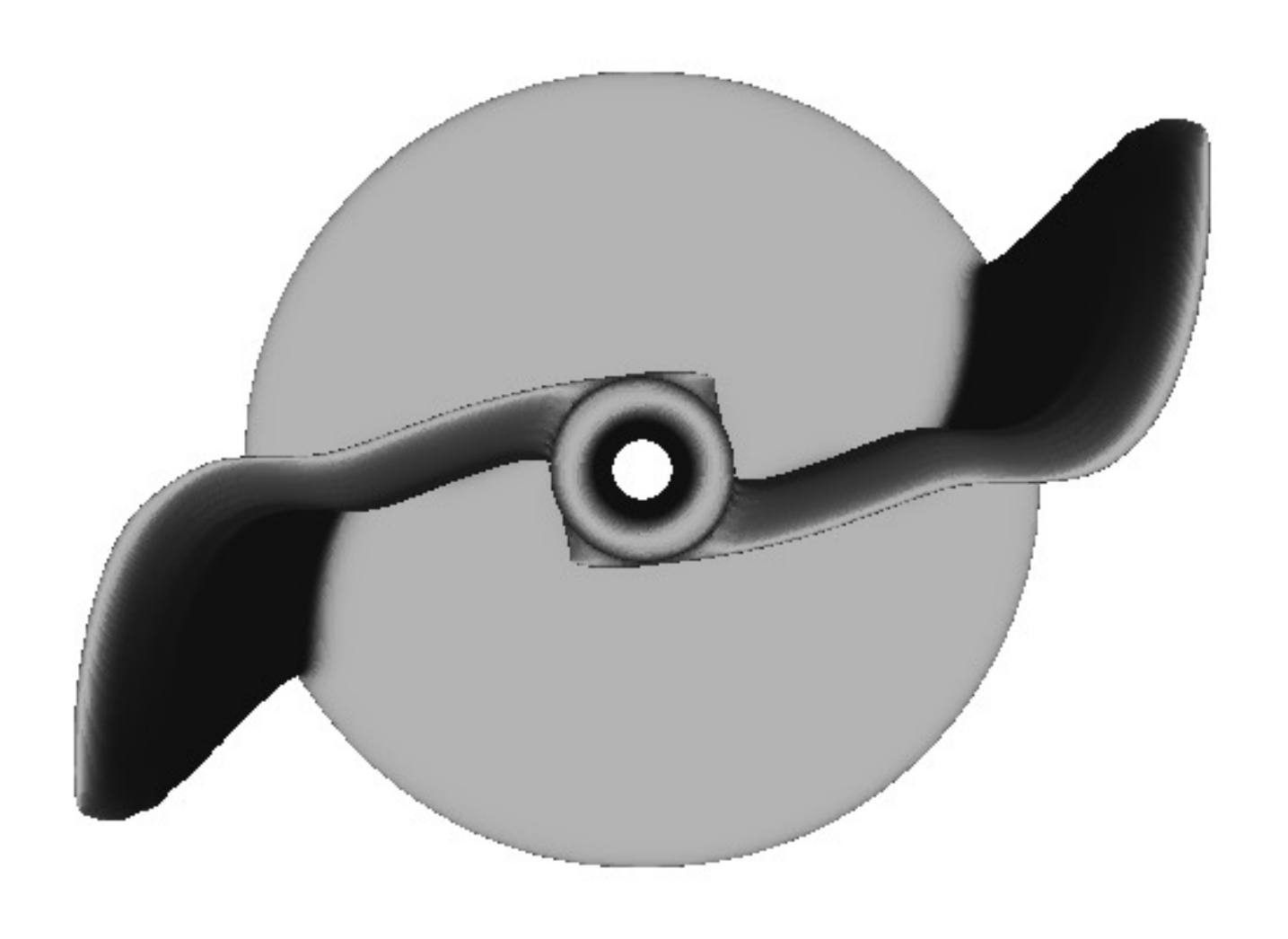}}
	\subfigure[s2]{\includegraphics[width=2.0cm]{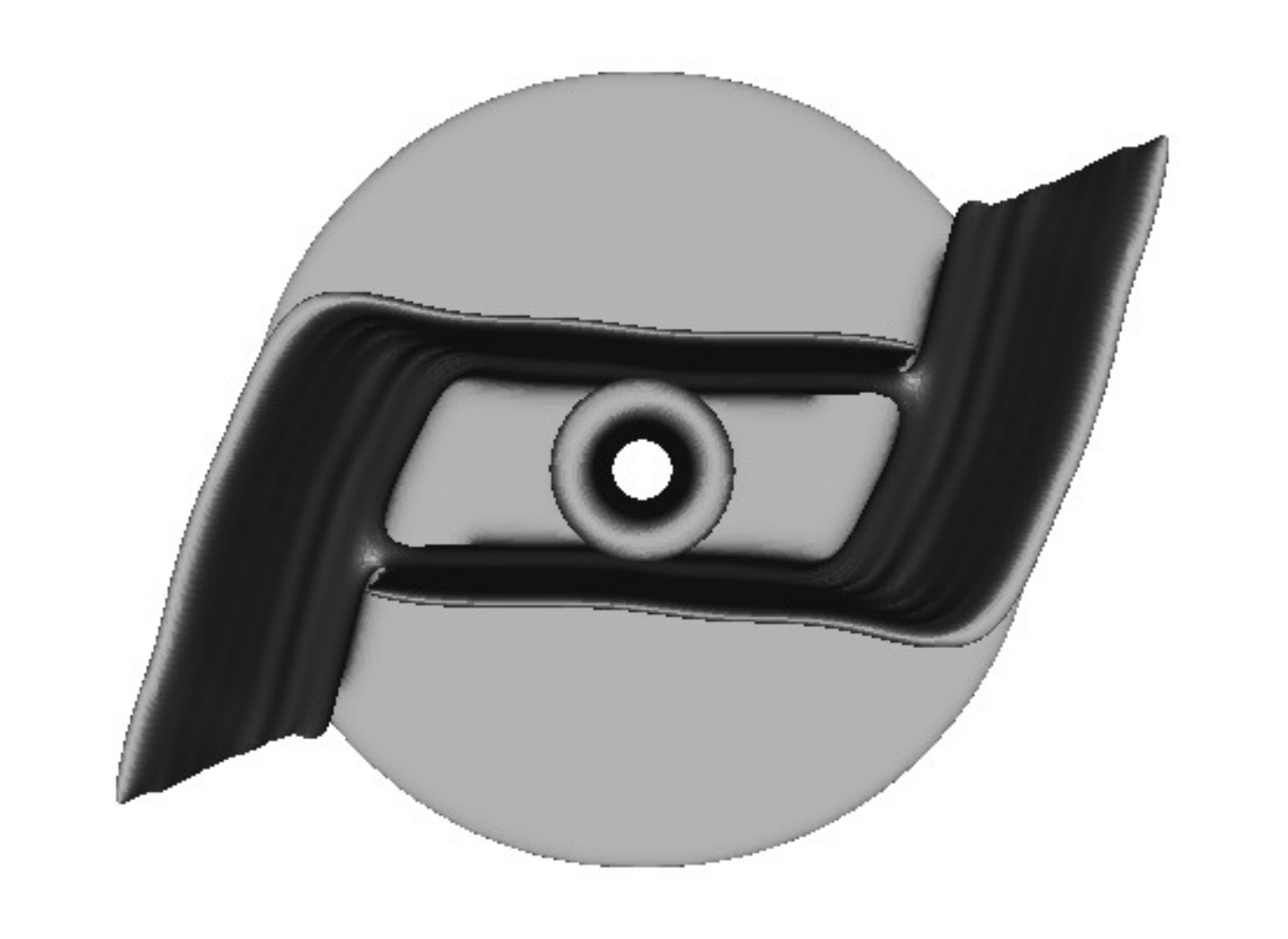}}
	\subfigure[s3]{\includegraphics[width=2.0cm]{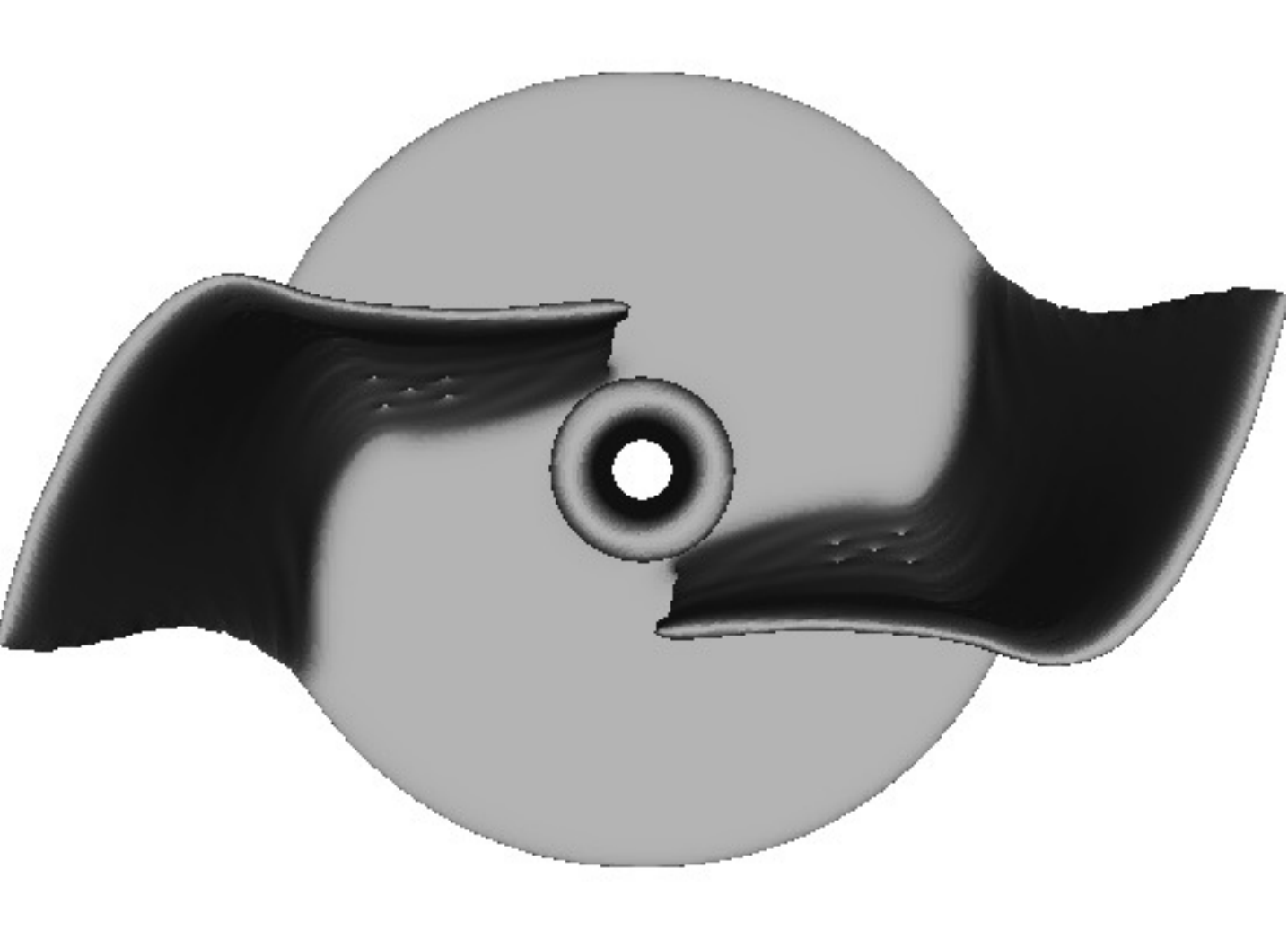}}
	\subfigure[s4]{\includegraphics[width=2.0cm]{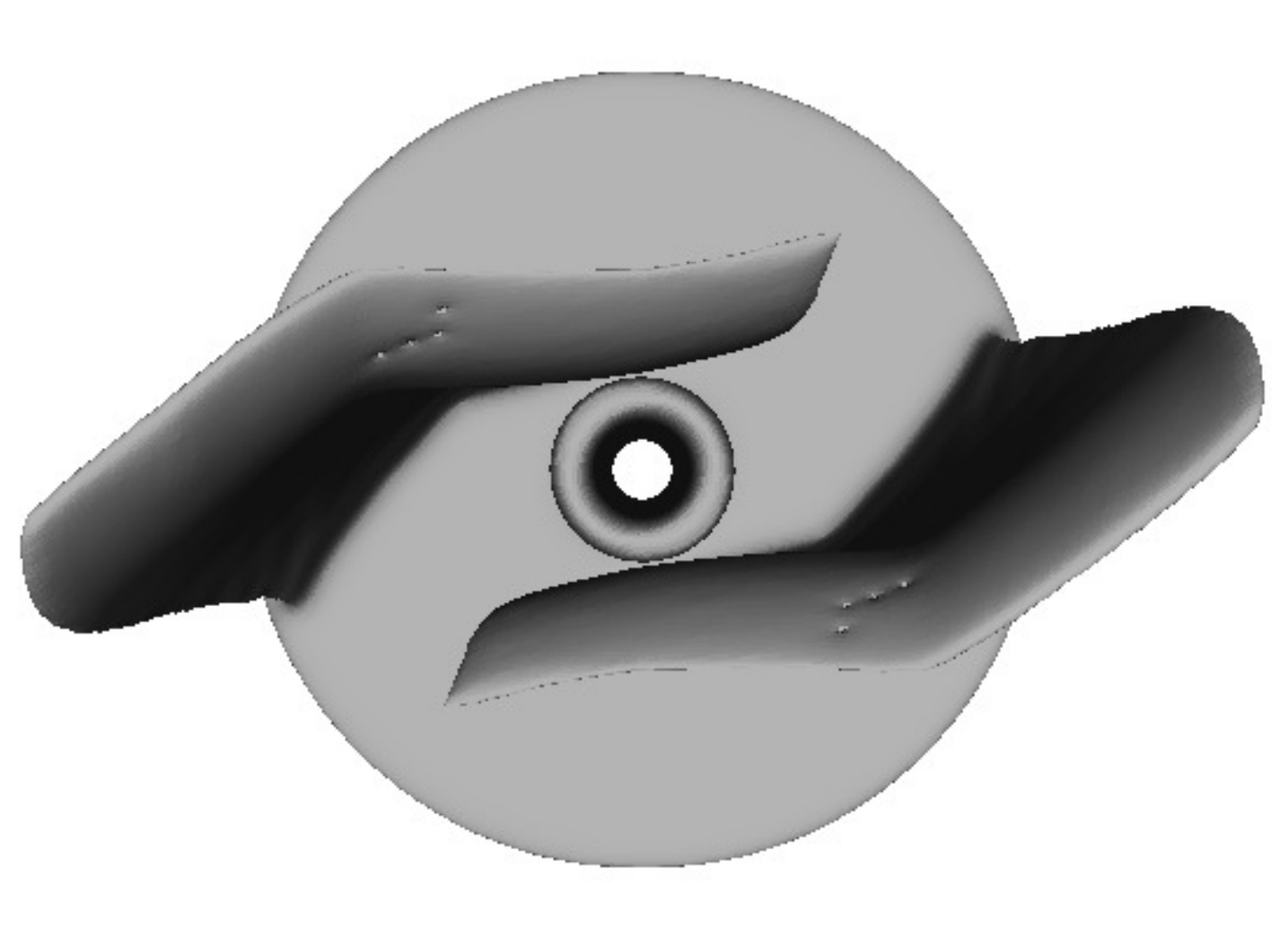}}
	\subfigure[s5]{\includegraphics[width=2.0cm]{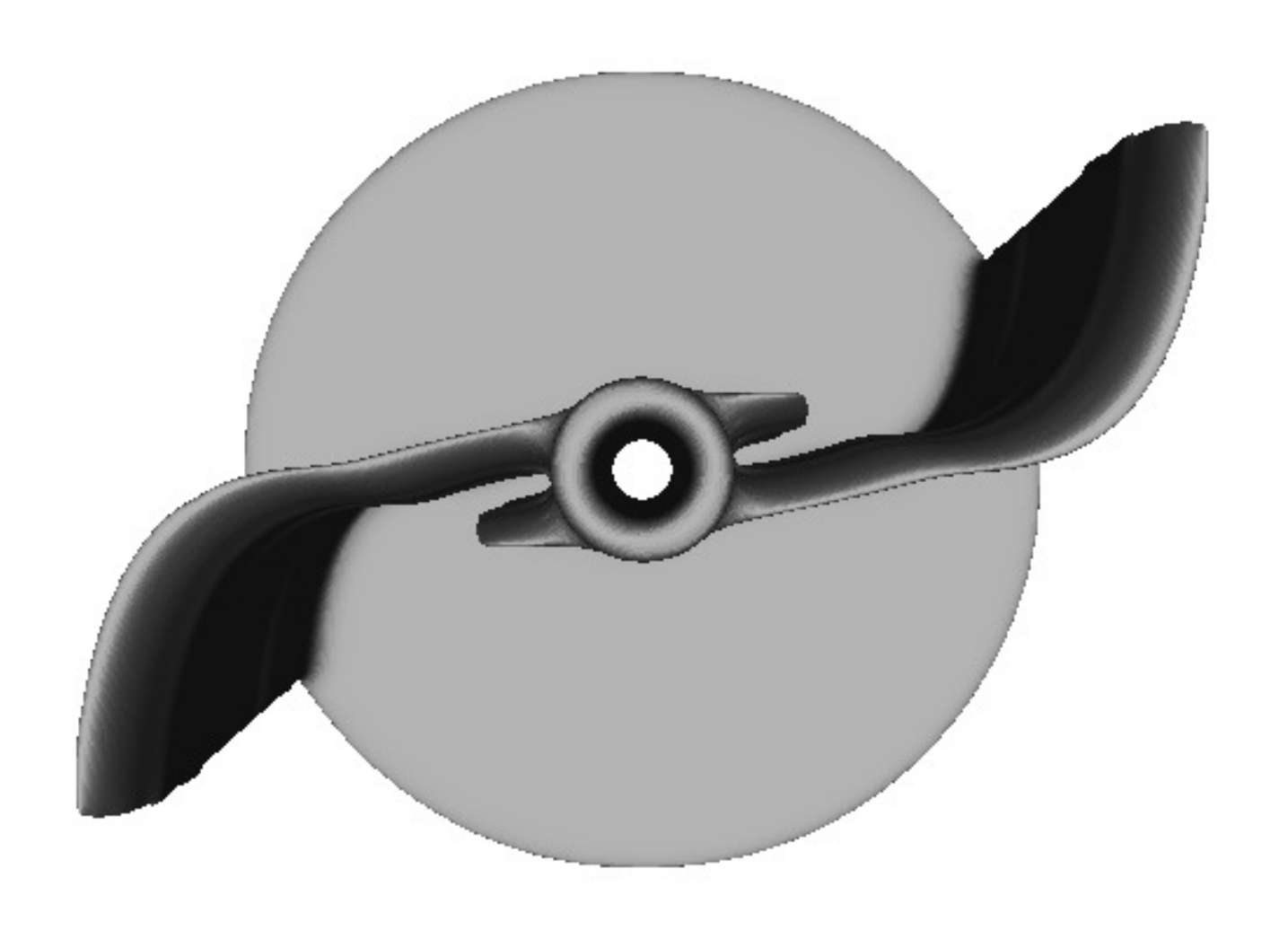}}
	\subfigure[s6]{\includegraphics[width=2.0cm]{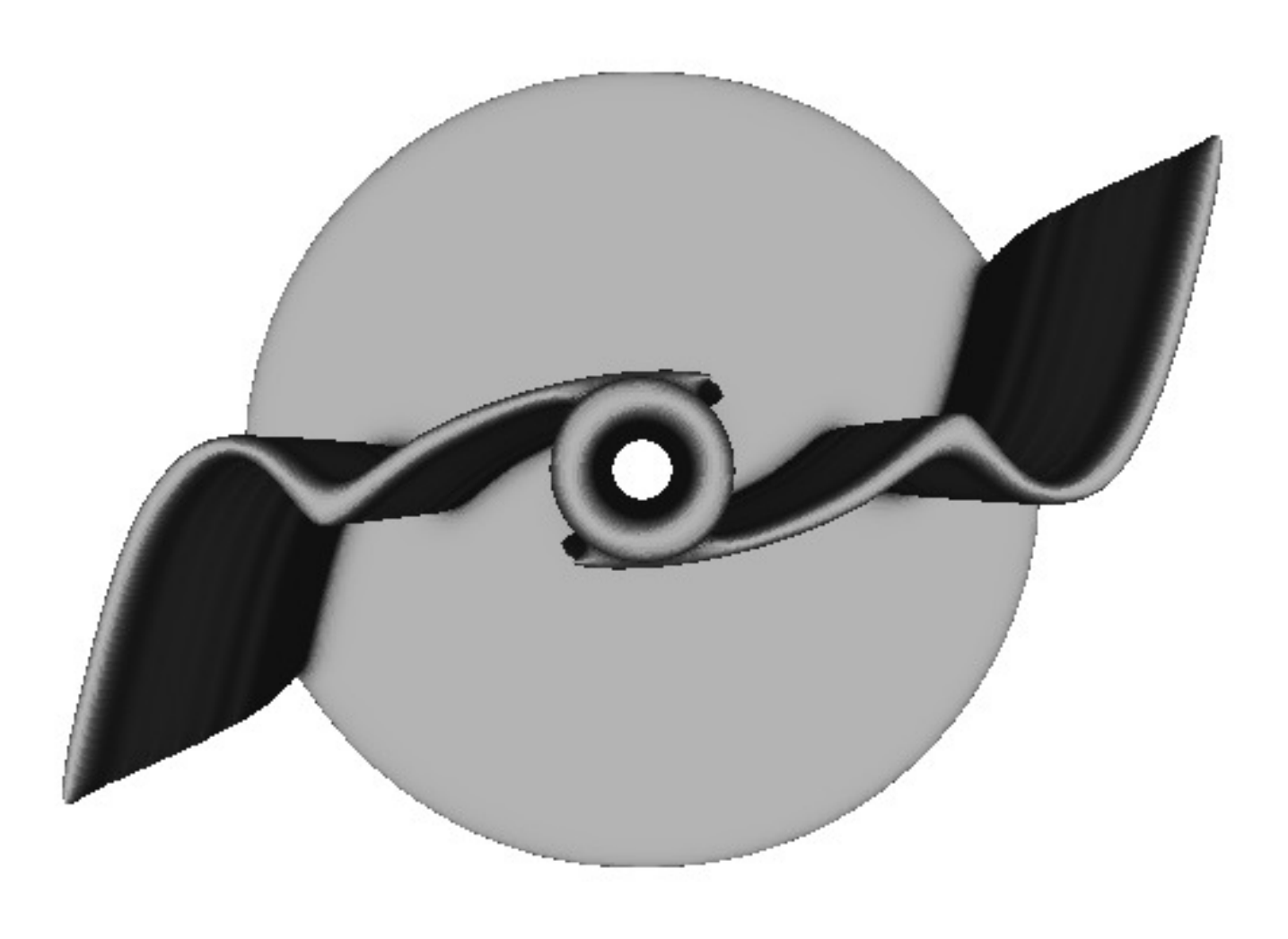}}
	\caption{Cross sections of the fittest evolved VAWT array after 2 CGA-b generations. Total $KE=7.6$~mJ, $m=44.8$~g, 2677~rpm.}
	\label{fig:cga-2gen}
\end{figure*}

\begin{figure*}[t]
	\centering 
	\subfigure[s1]{\includegraphics[width=2.0cm]{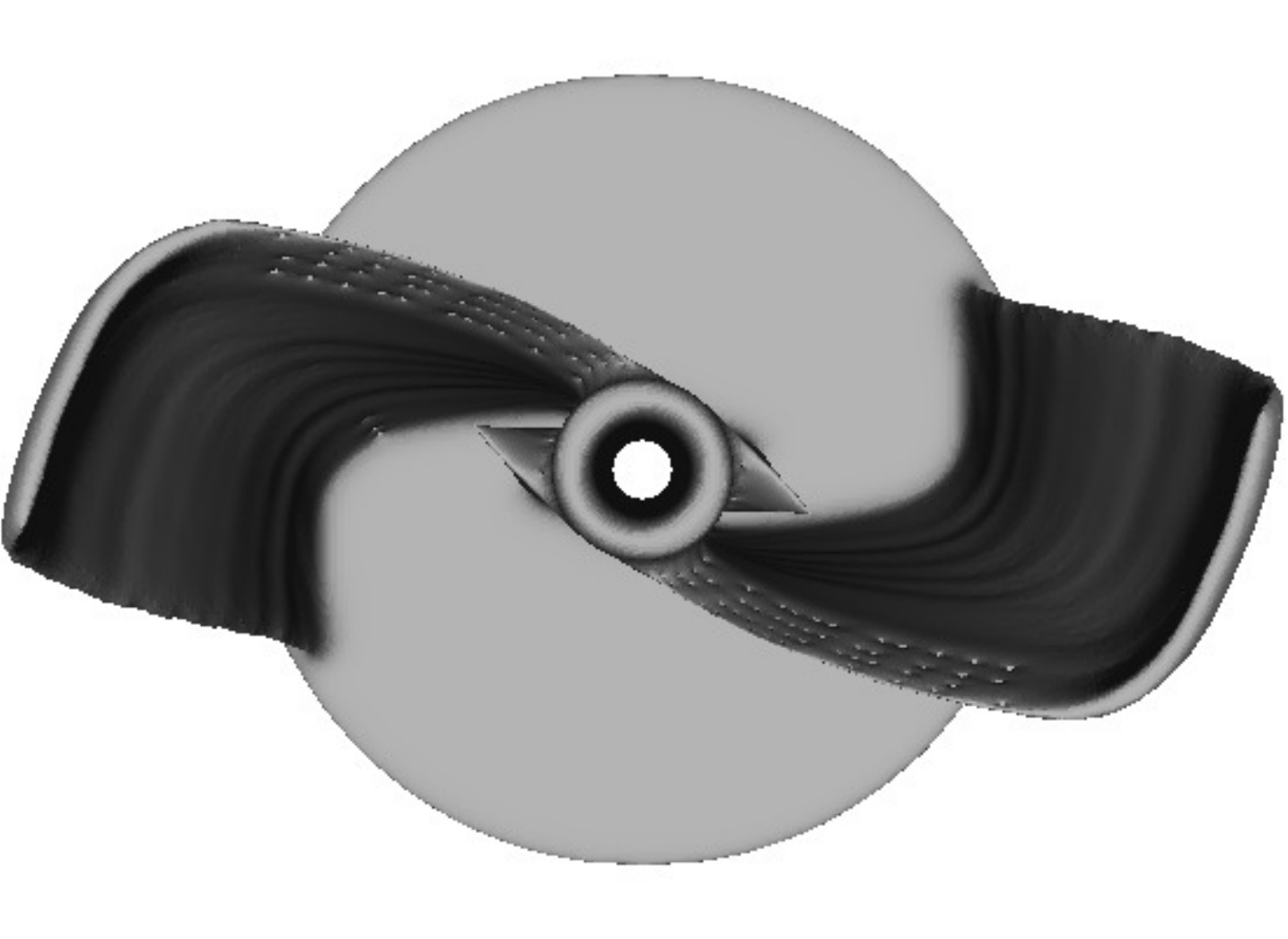}}
	\subfigure[s2]{\includegraphics[width=2.0cm]{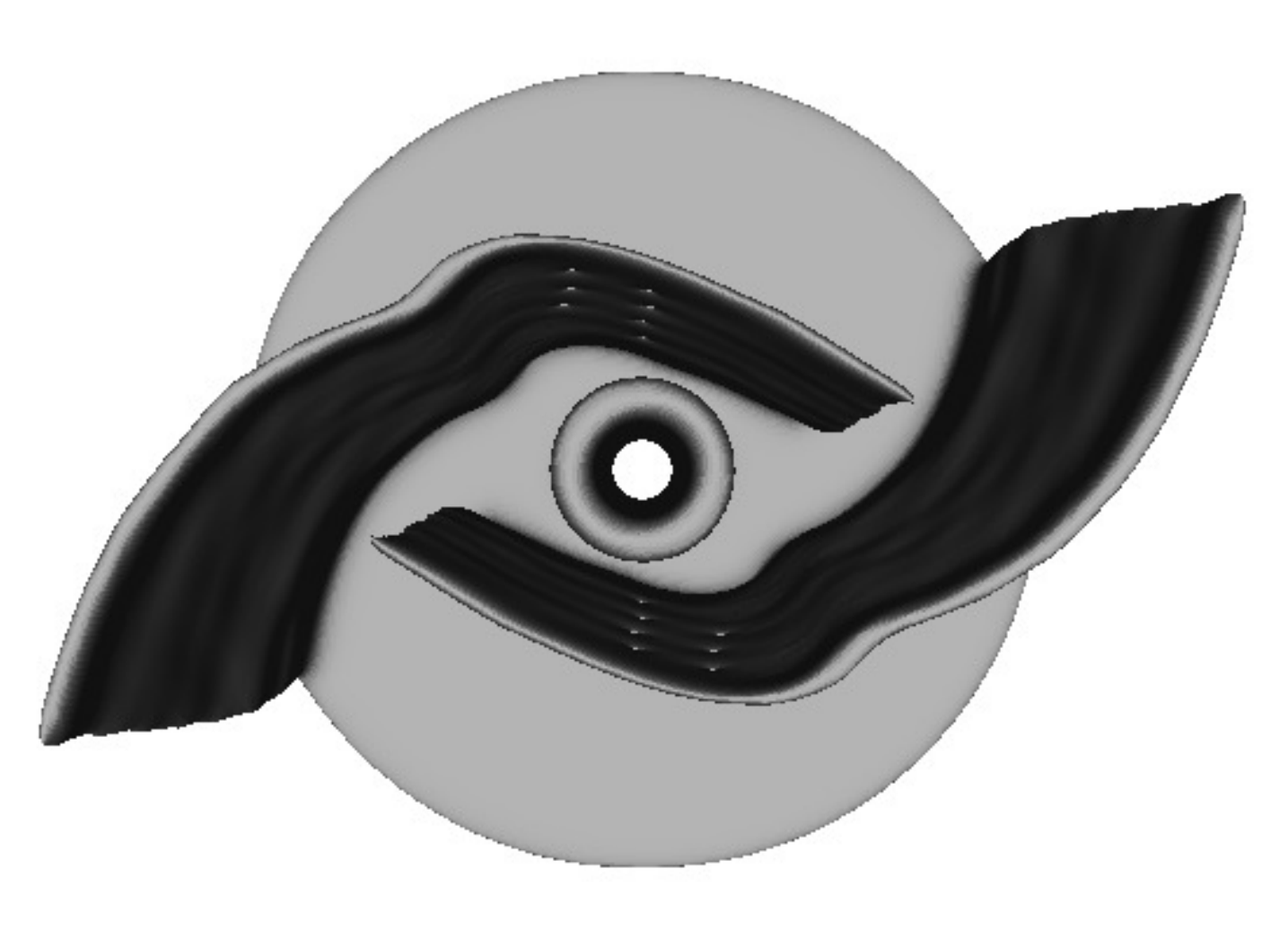}\label{cga2}}
	\subfigure[s3]{\includegraphics[width=2.0cm]{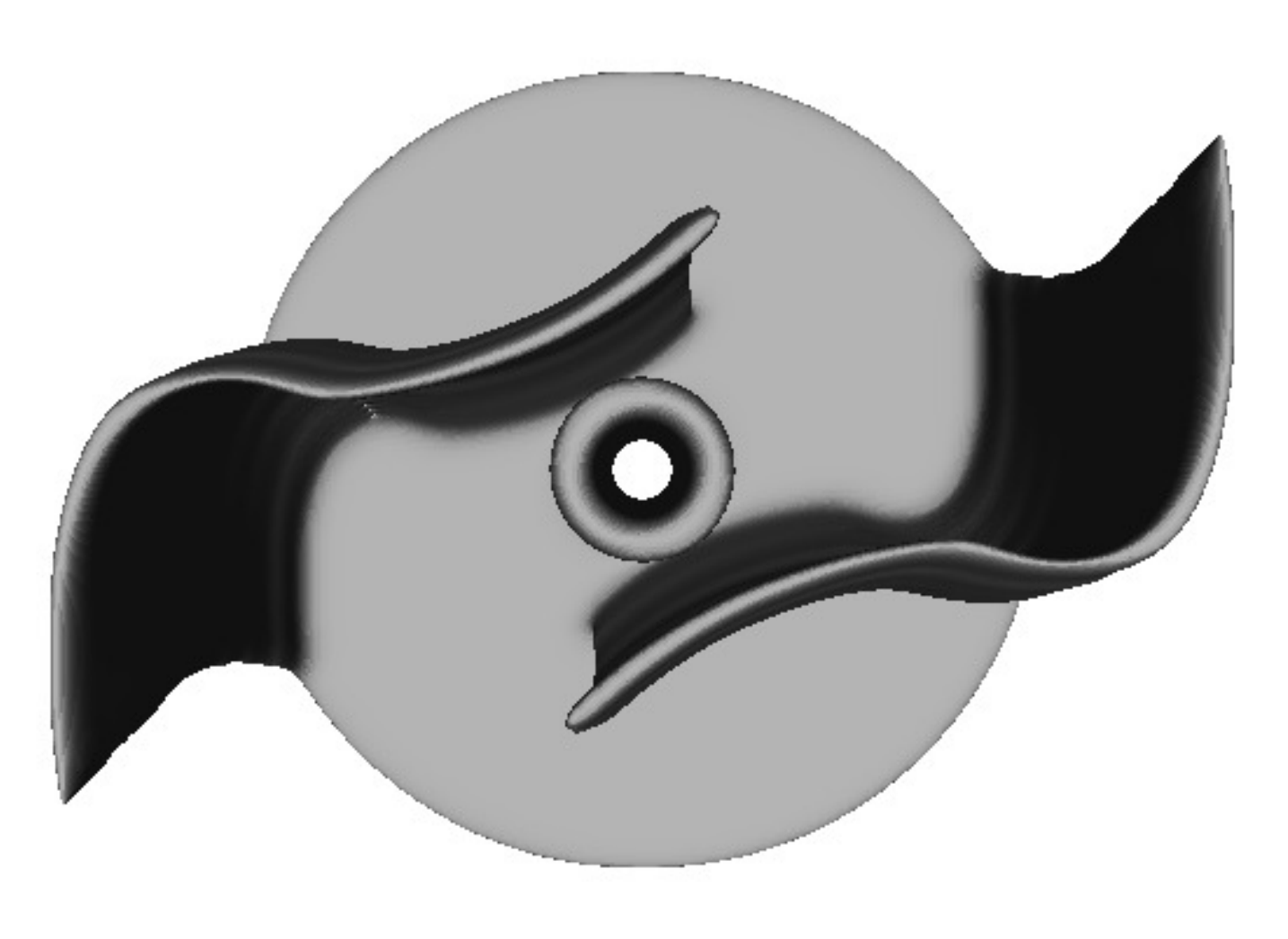}}
	\subfigure[s4]{\includegraphics[width=2.0cm]{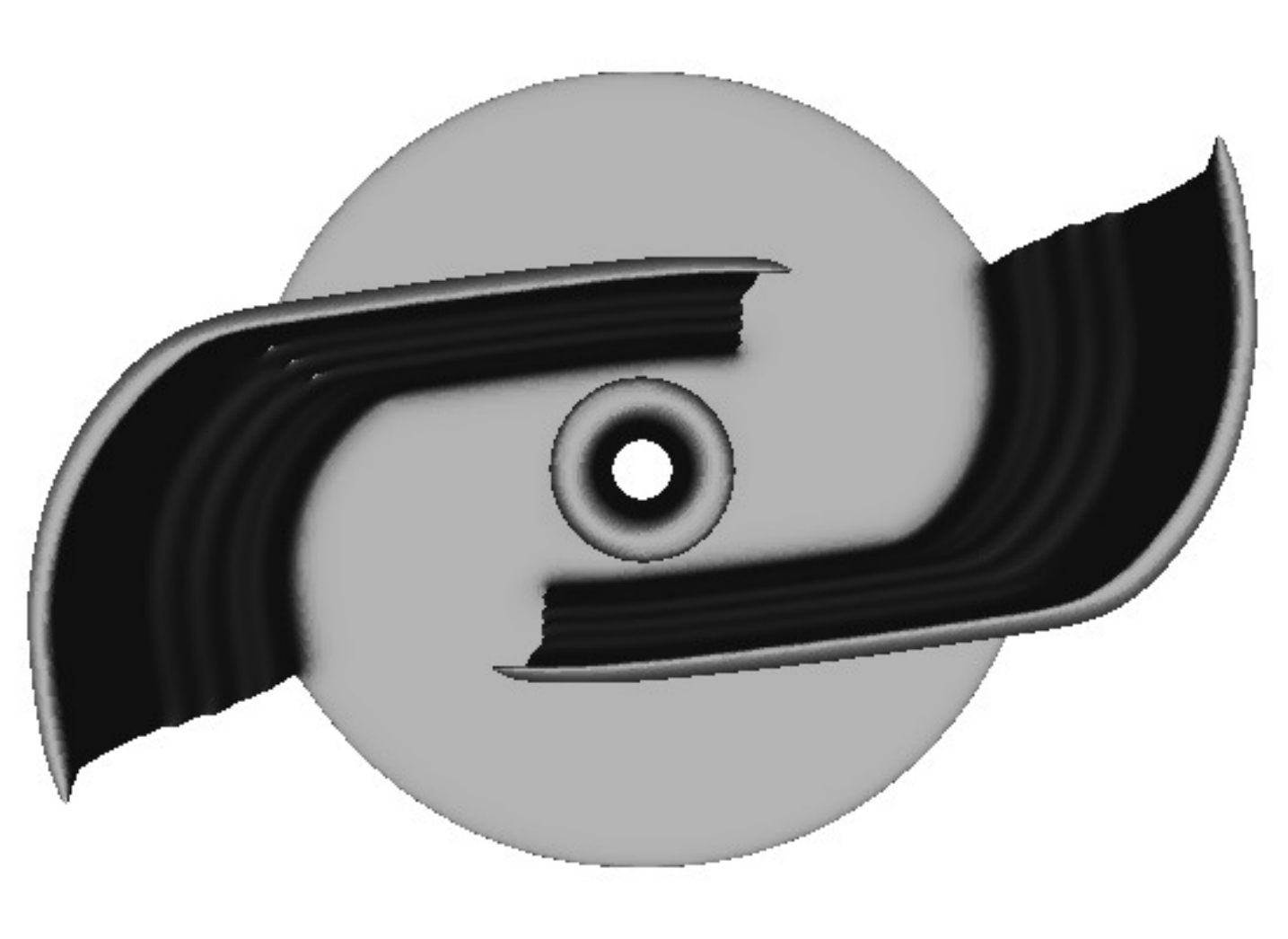}\label{cga4}}
	\subfigure[s5]{\includegraphics[width=2.0cm]{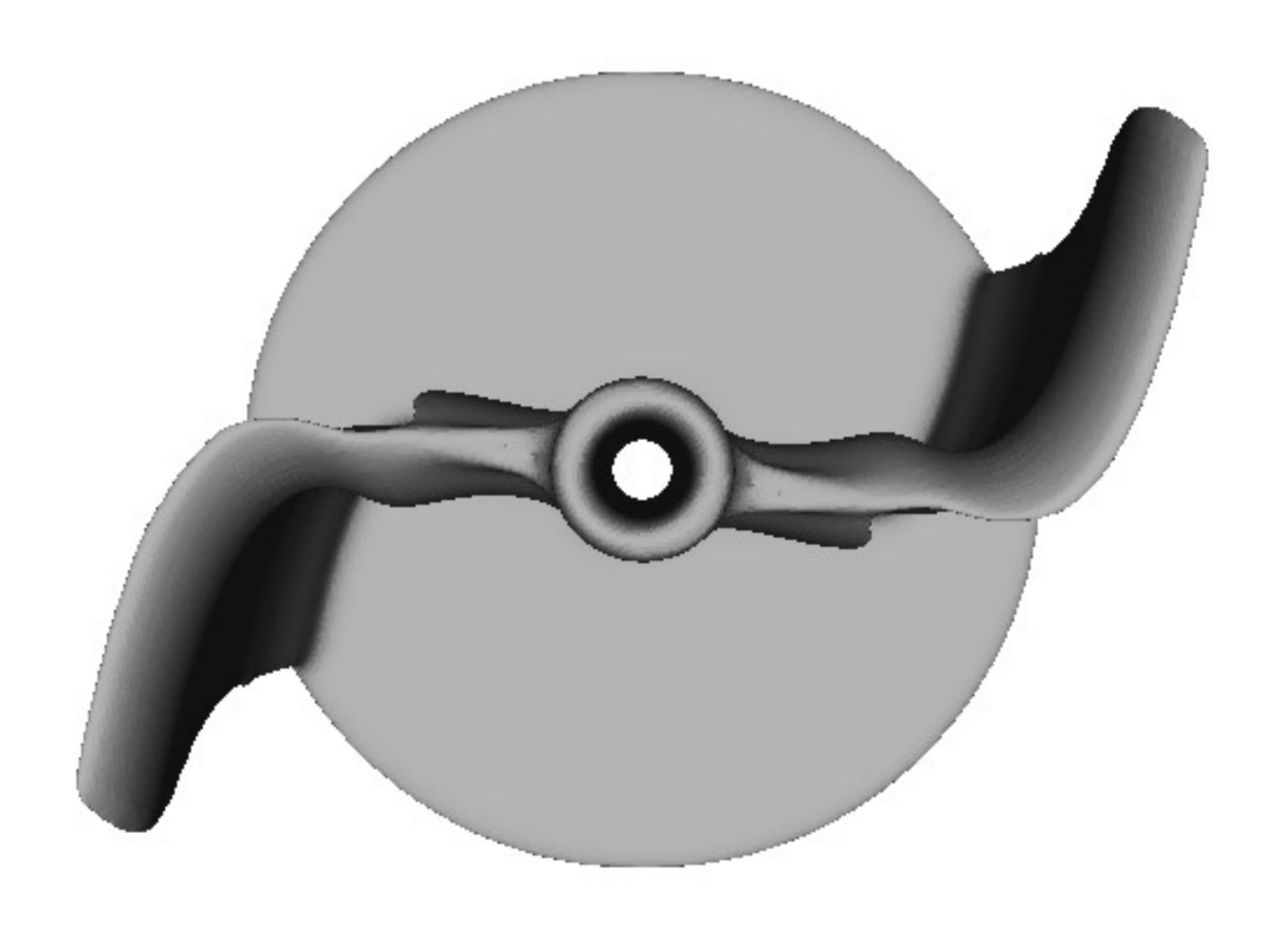}}
	\subfigure[s6]{\includegraphics[width=2.0cm]{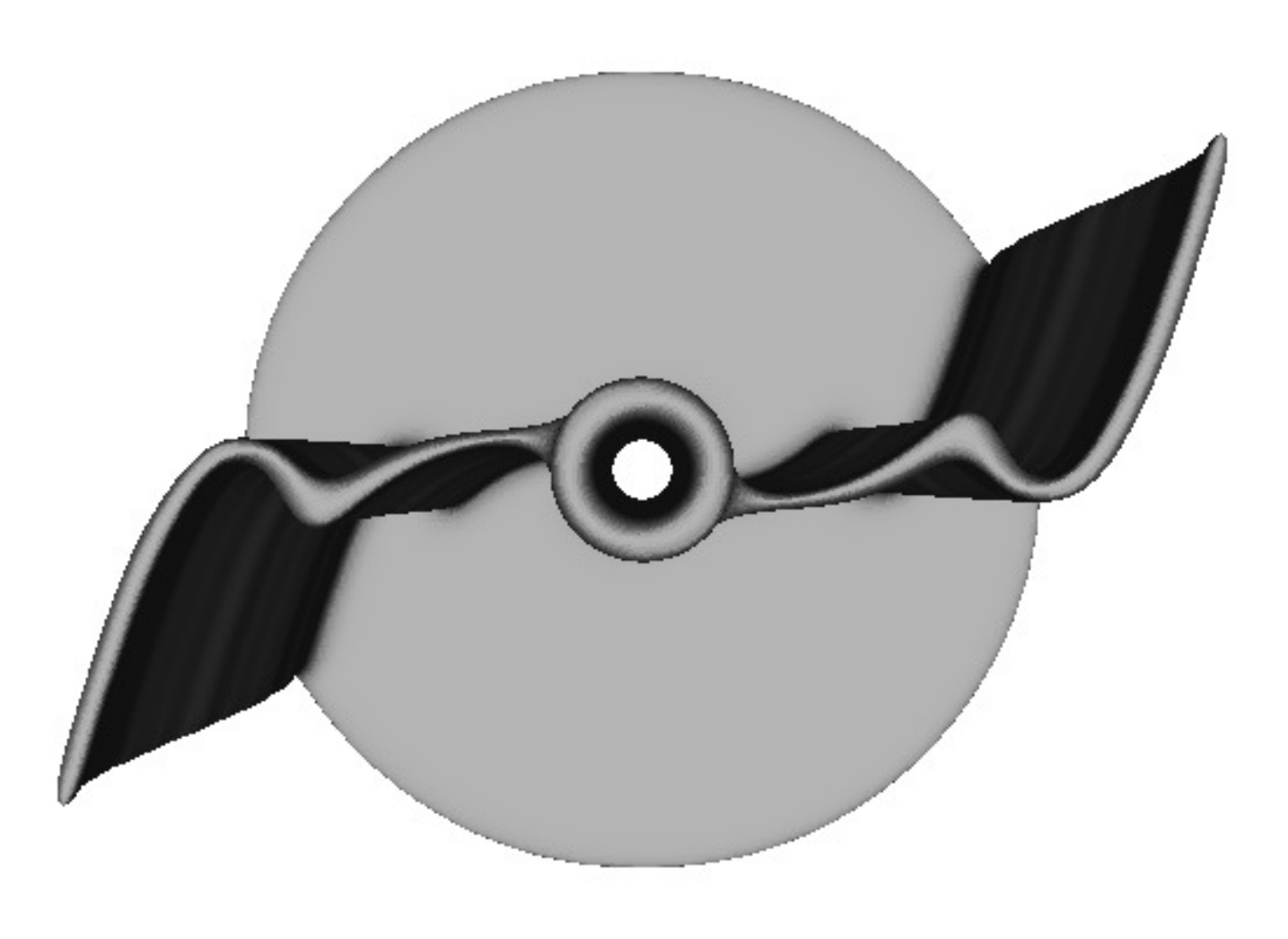}}
	\caption{Cross sections of the fittest evolved VAWT array after 3 CGA-b generations. Total $KE=10$~mJ, $m=45.6$~g, 3004~rpm.}
	\label{fig:cga-3gen}
\end{figure*}

\begin{figure*}[t]
	\centering 
	\subfigure[s1]{\includegraphics[width=2.0cm]{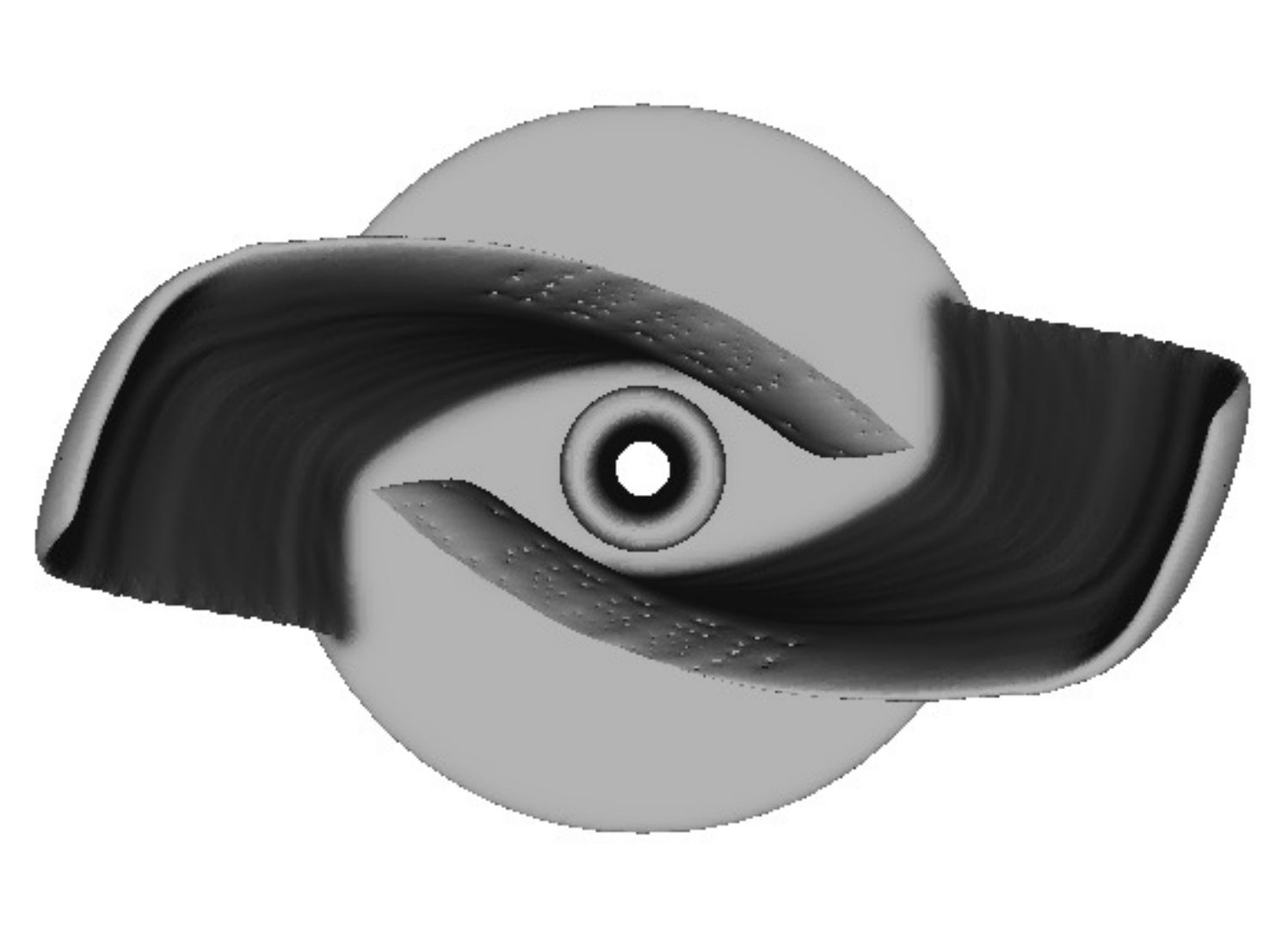}}
	\subfigure[s2]{\includegraphics[width=2.0cm]{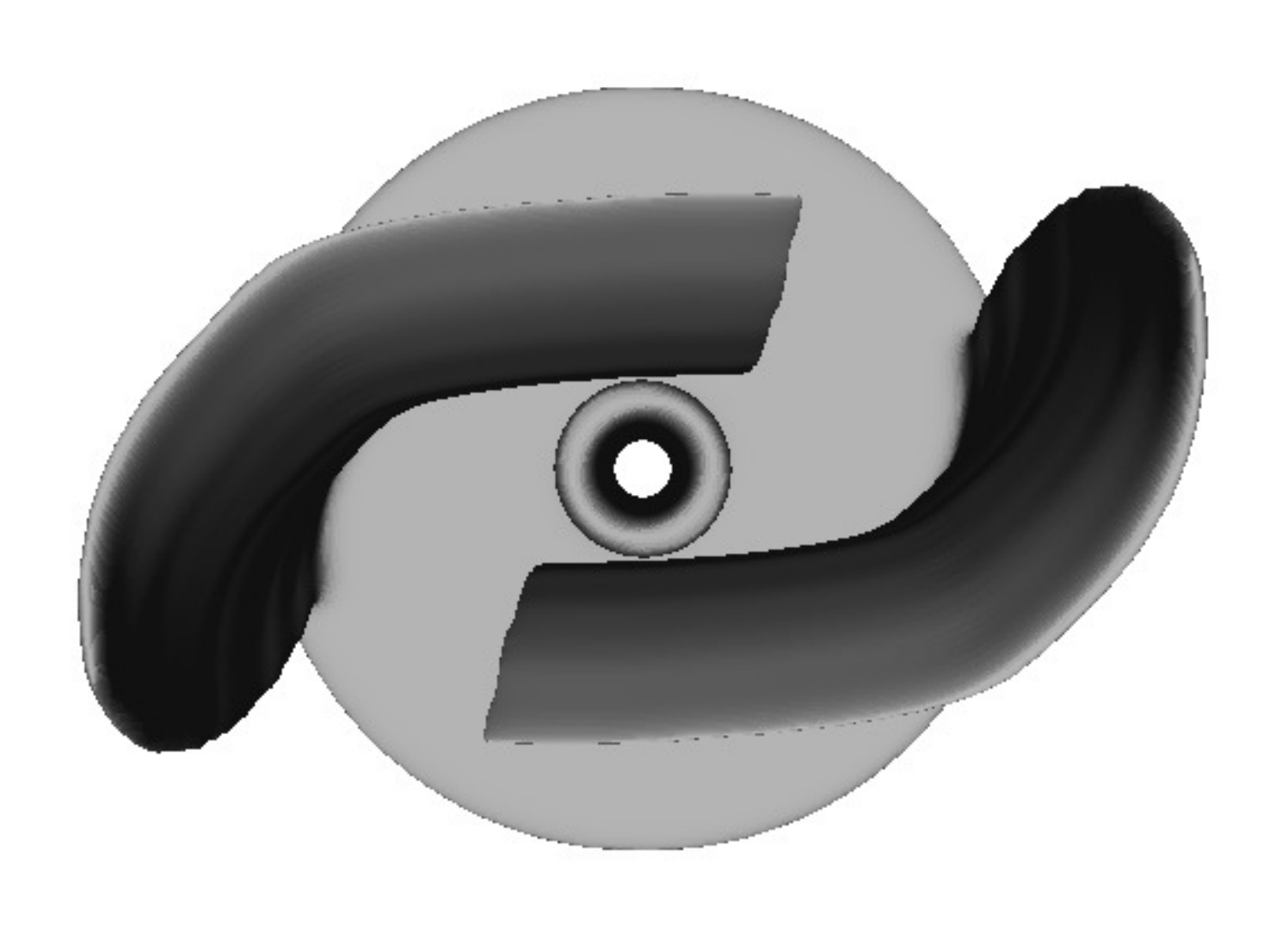}}
	\subfigure[s3]{\includegraphics[width=2.0cm]{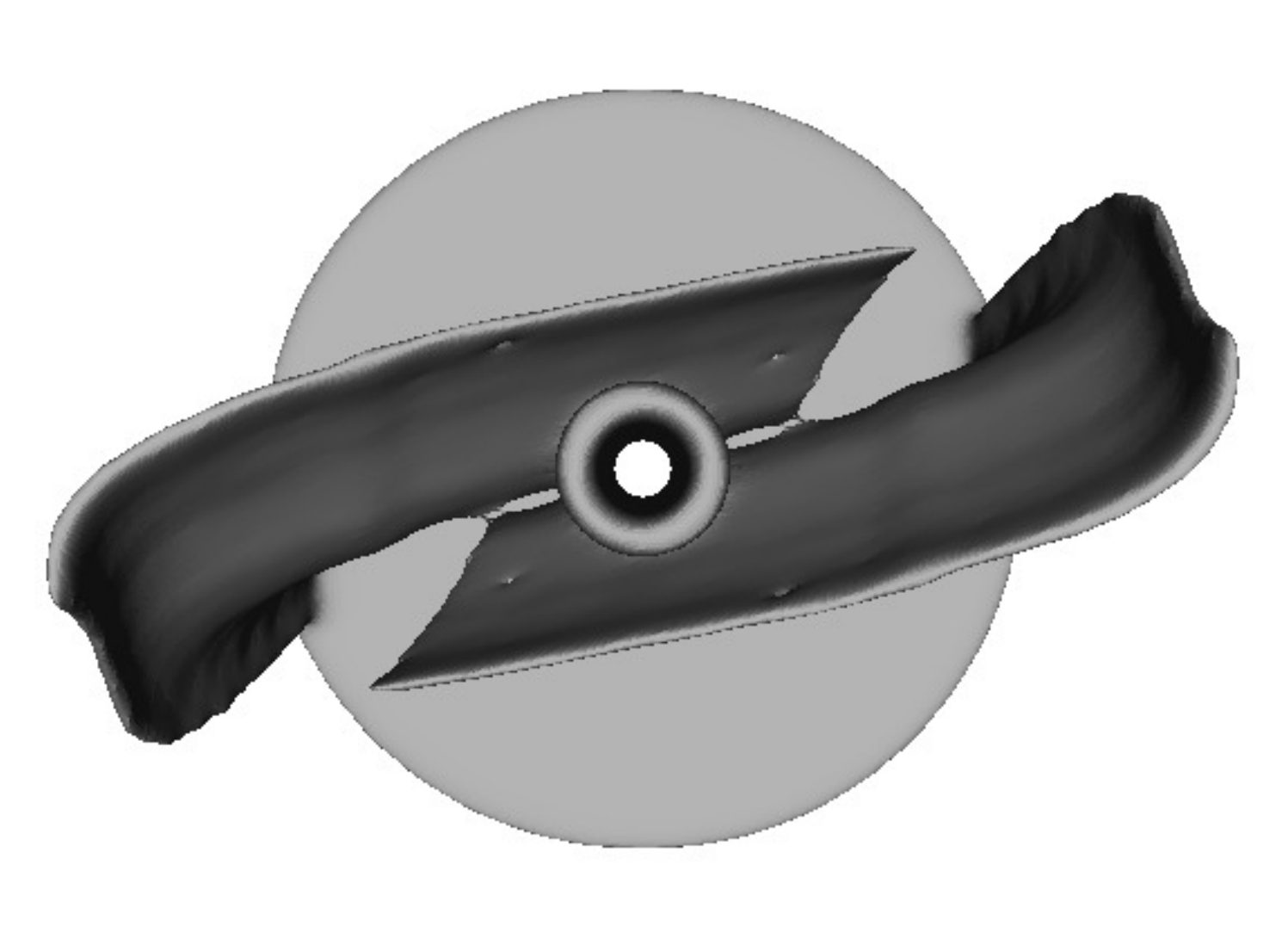}}
	\subfigure[s4]{\includegraphics[width=2.0cm]{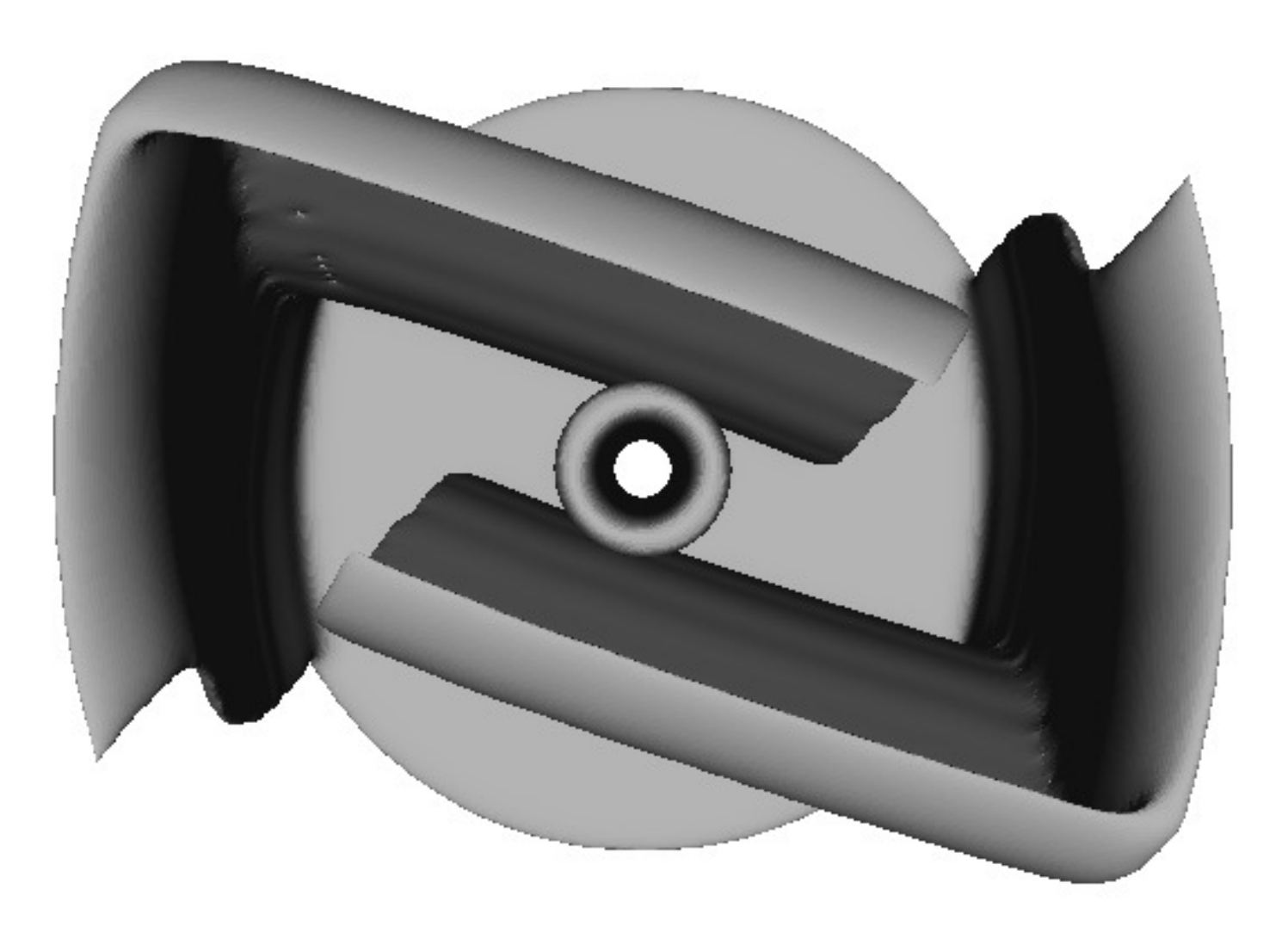}}
	\subfigure[s5]{\includegraphics[width=2.0cm]{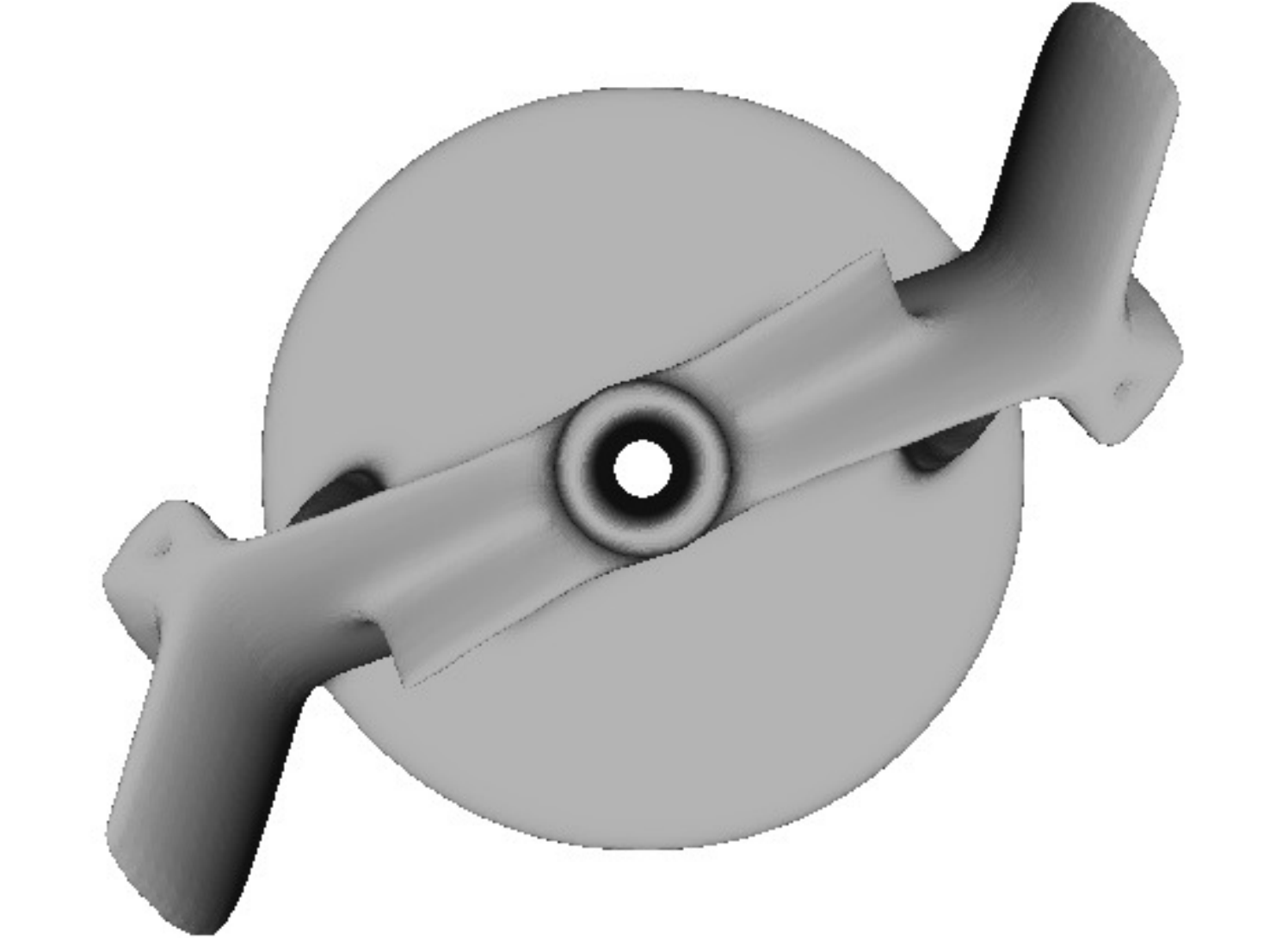}}
	\subfigure[s6]{\includegraphics[width=2.0cm]{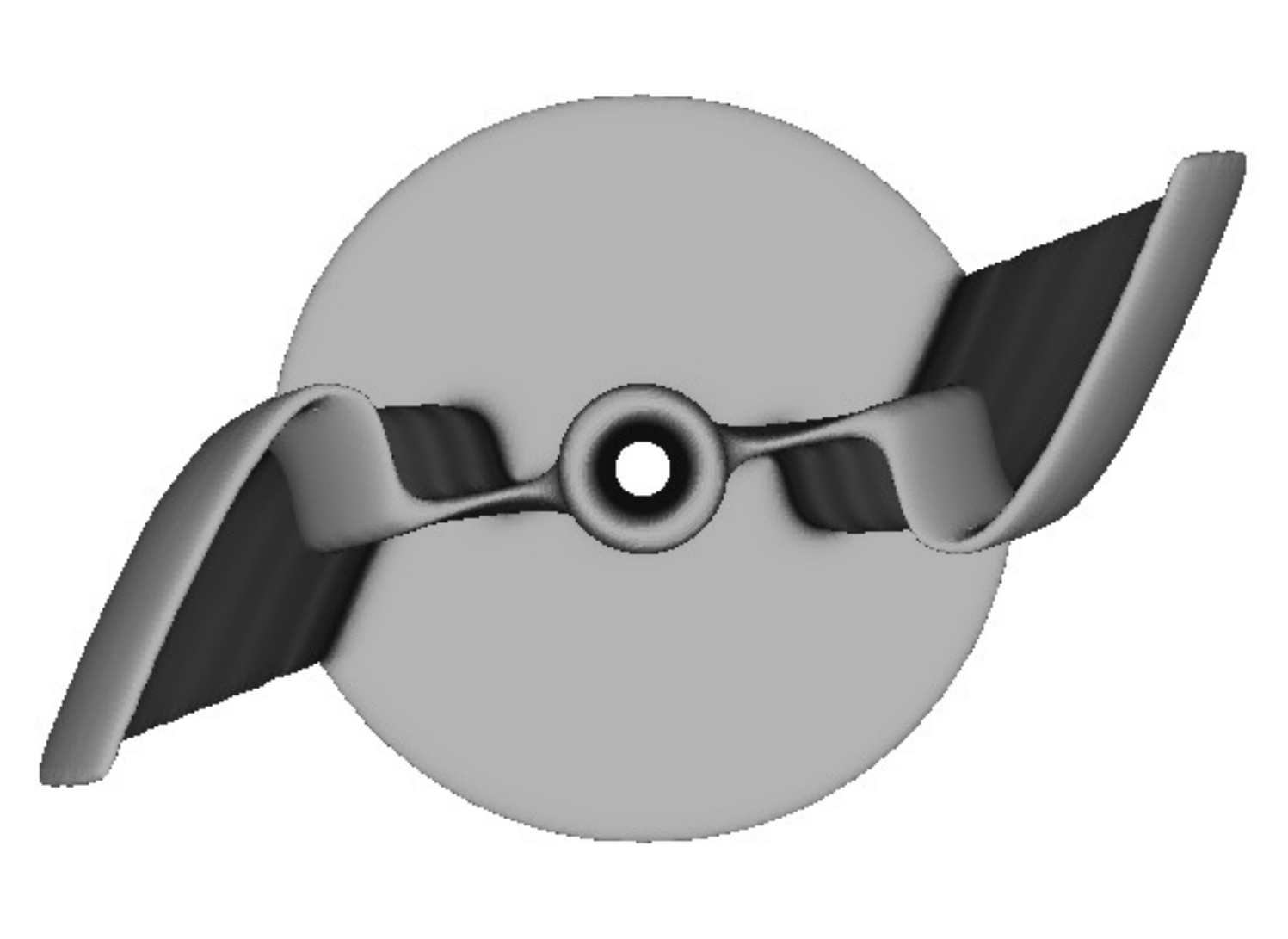}}
	\caption{Cross sections of the fittest evolved VAWT array after 3 CGA-b generations plus 1 SCGA-b generation. Total $KE=12.2$~mJ, $m=49.3$~g, 3094~rpm.}
	\label{fig:scga-gen}
\end{figure*}

\begin{figure*}[t]
	\centering 
	\subfigure[s1]{\includegraphics[width=2.5cm]{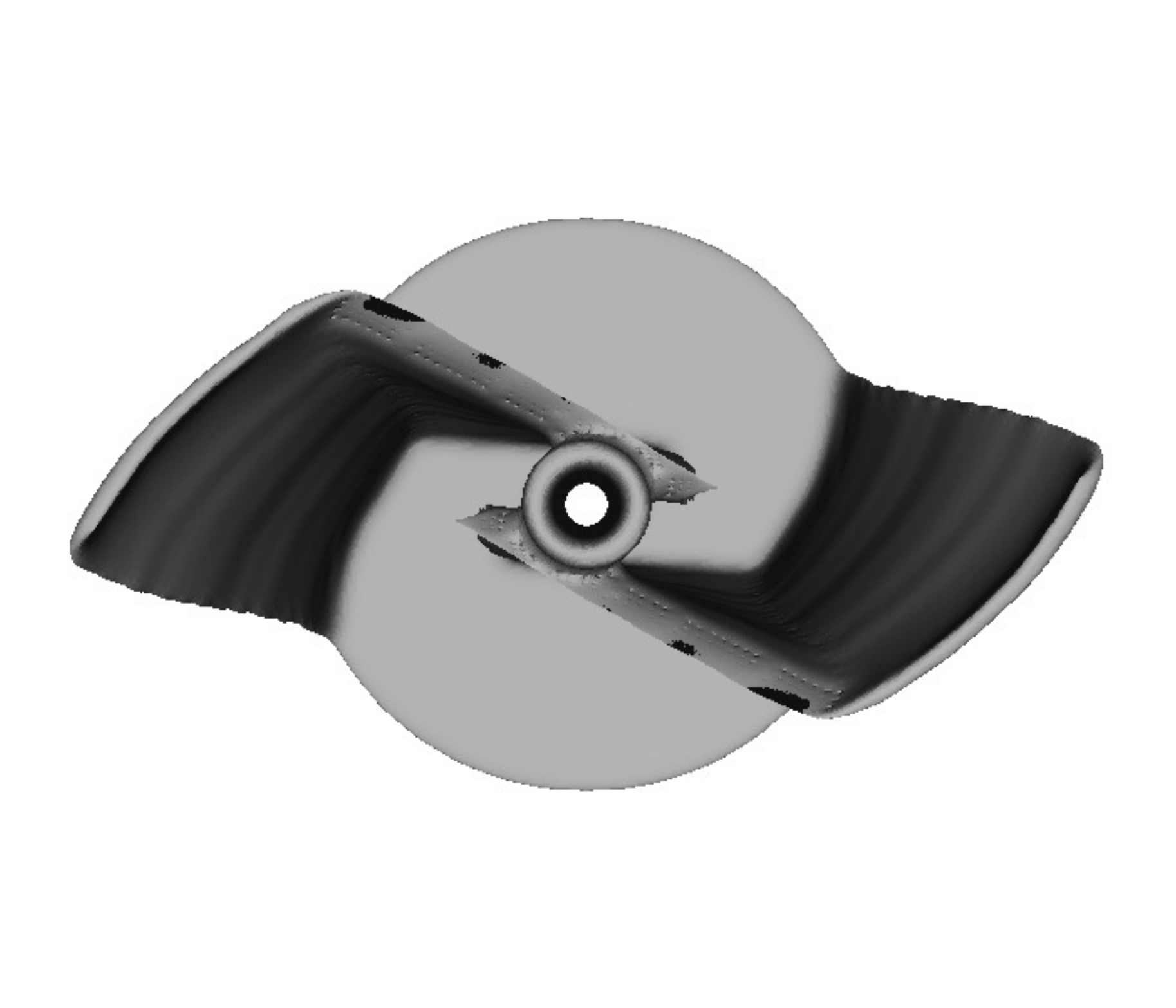}} \hspace{-4mm}
	\subfigure[s2]{\includegraphics[width=2.5cm]{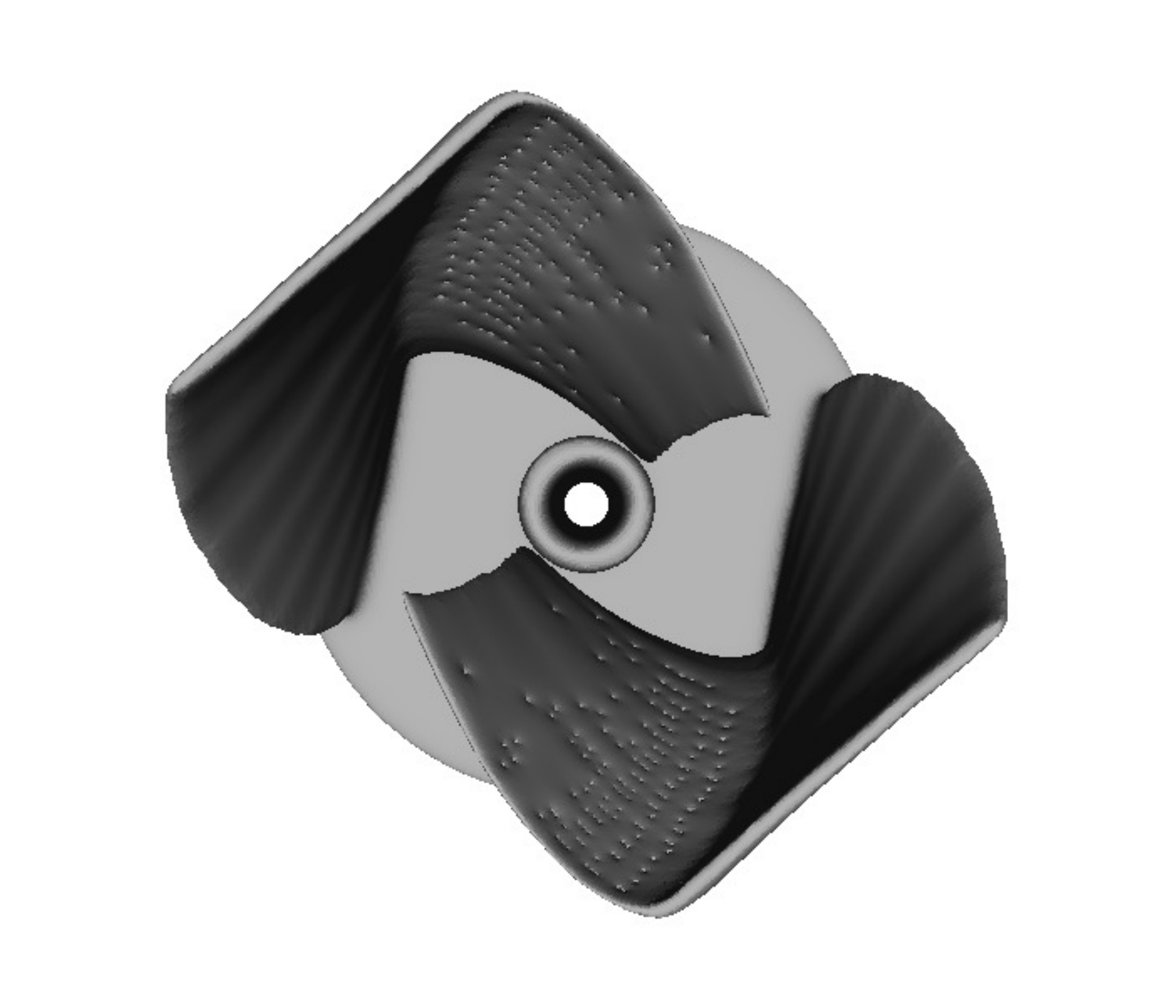}\label{ws2}} \hspace{-4mm}
	\subfigure[s3]{\includegraphics[width=2.5cm]{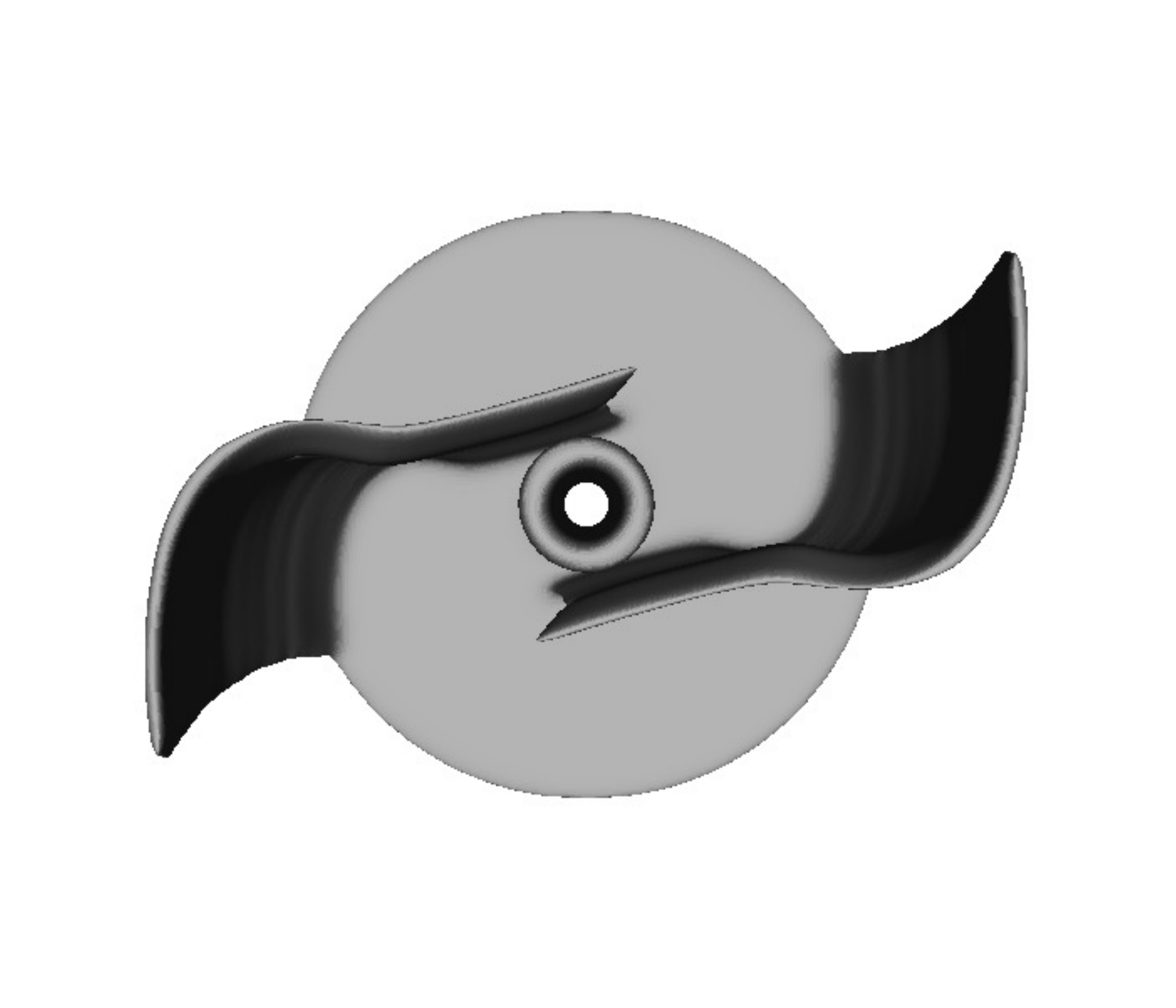}} \hspace{-4mm}
	\subfigure[s4]{\includegraphics[width=2.5cm]{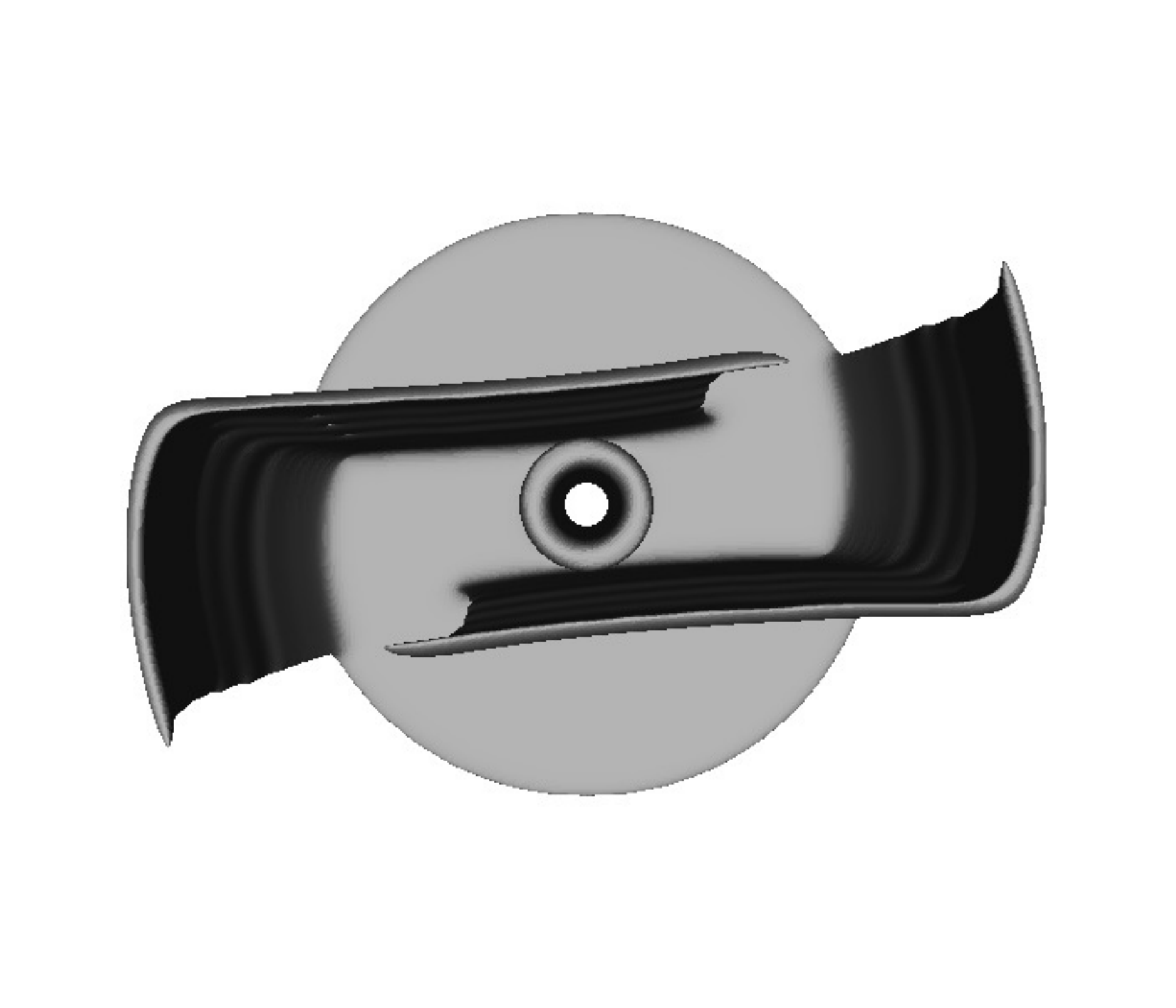}\label{ws4}} \hspace{-4mm}
	\subfigure[s5]{\includegraphics[width=2.5cm]{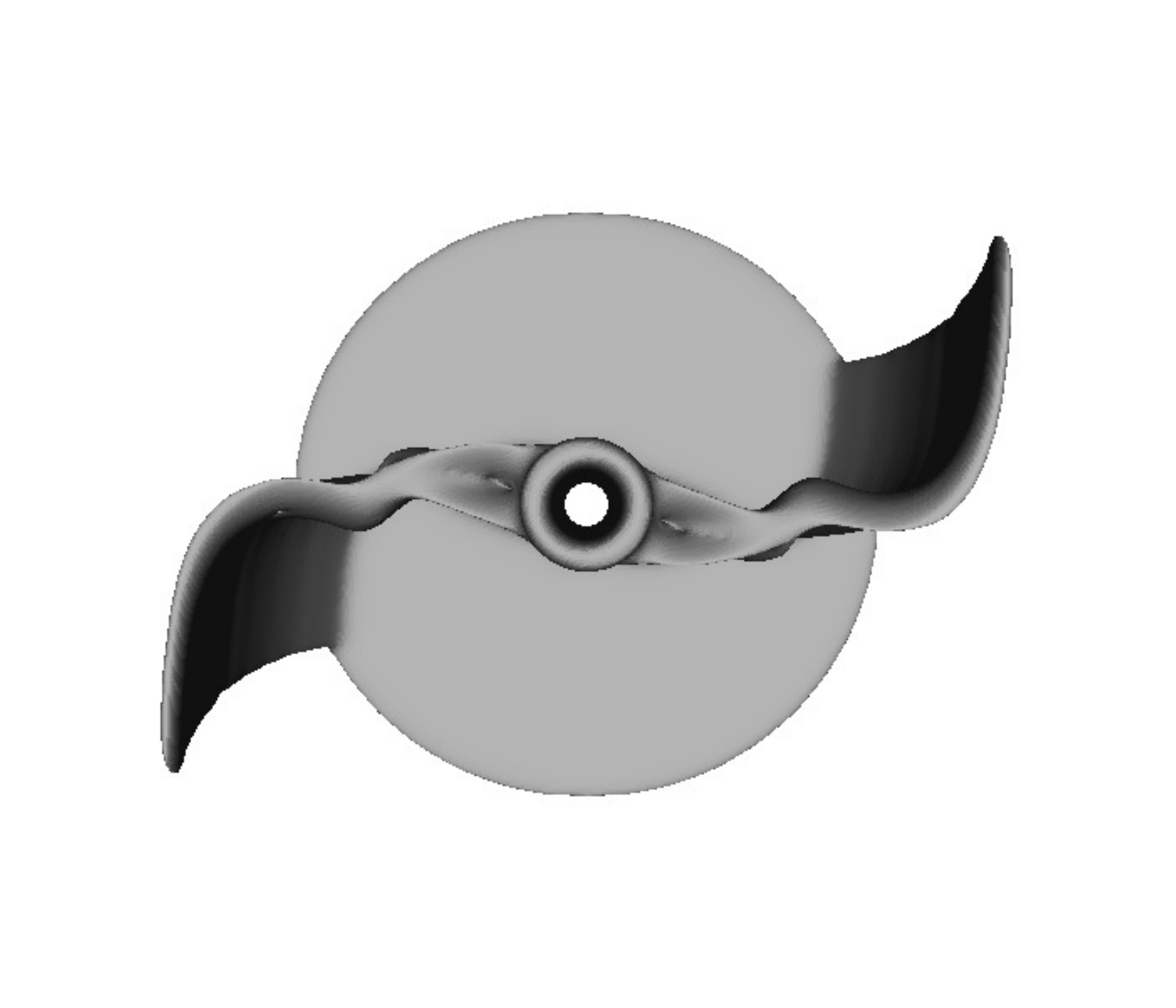}} \hspace{-4mm}
	\subfigure[s6]{\includegraphics[width=2.5cm]{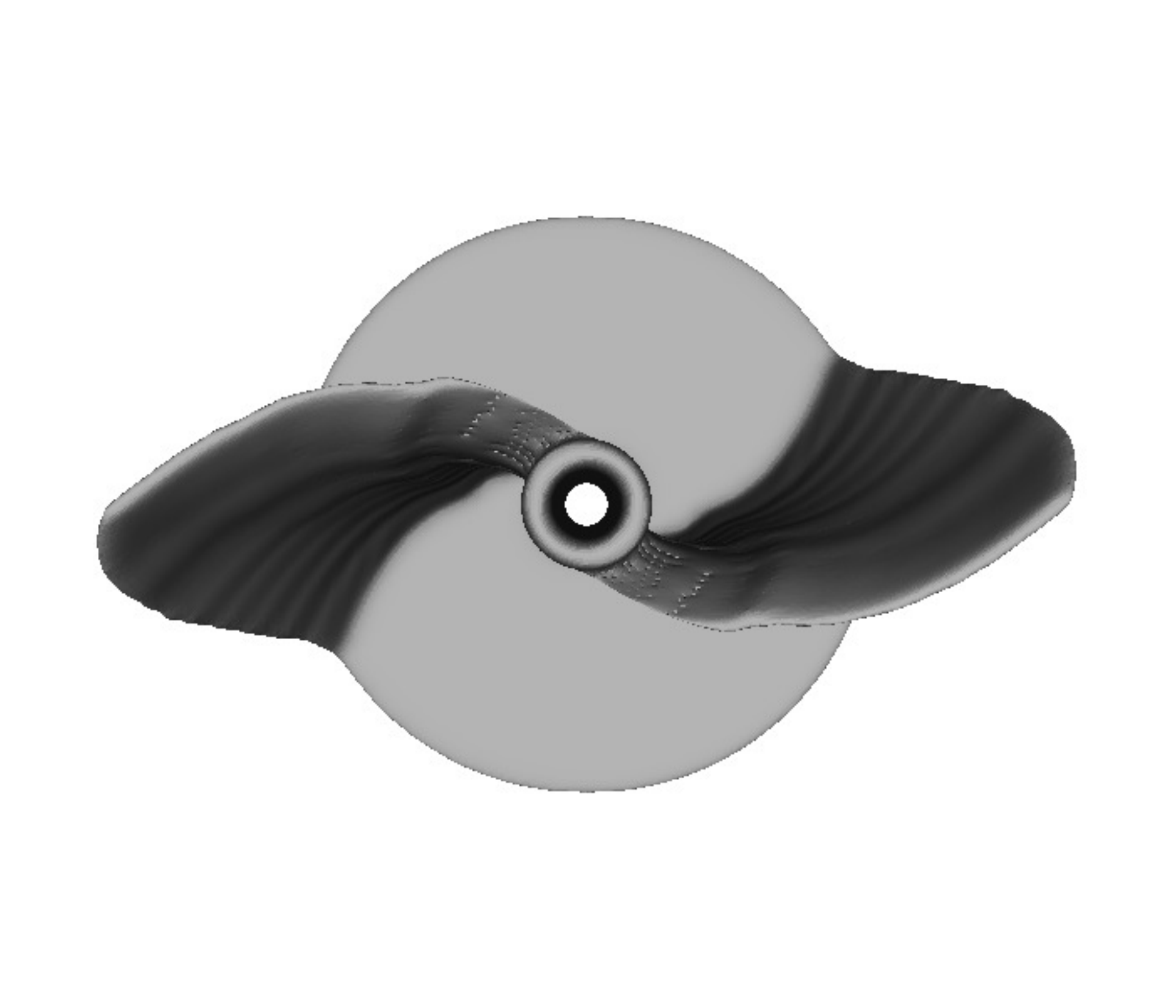}}
	\caption{Cross sections of the fittest evolved VAWT array after 3 CGA-b generations plus 1 SCGA-bw generation. Total $KE=14.8$~mJ, $m=45.9$~g, 3668~rpm.}
	\label{fig:scgaw-gen}
\end{figure*}

Figure~\ref{fig:ke} shows the total angular kinetic energy of the fittest arrays each generation; Figure~\ref{fig:mass} shows the total mass, and Figure~\ref{fig:rpm} the total rpm. The cross sections of the fittest array designs can be seen in Figures~\ref{fig:cga-1gen}--\ref{fig:scgaw-gen}. When the position of the final evolved SCGA-b array was inverted, i.e., the first species design being swapped with the sixth, second swapped with fifth, etc., a decrease in total rpm of 17.8\% was observed, causing a reduction in total $KE$ of 36.7\%. A similar test was performed for the final evolved SCGA-bw array and the total rpm decreased by 14\% with a consequential decrease in total $KE$ of 22\%, showing that evolution has exploited position-specific characteristics.
     
It is interesting to note the similarity of some of the evolved VAWTs with human engineered designs. Bach~\cite{Bach:1931} performed one of the earliest morphological studies of Savonius VAWT and found increased aerodynamic performance with a blade profile consisting of a $2/3$ flattened trailing section and a larger blade overlap to reduce the effect of the central shaft, which is similar to the fourth species designs in Figures~\ref{cga4} and~\ref{ws4}. The evolved VAWT in the second species, e.g.,\ Figures~\ref{cga2} and~\ref{ws2}, overall appear more rounded and similar to the classic S-shape Savonius design~\cite{Savonius:1930}. There appears to be little twist rotation along the $z$-axis of the evolved designs, which may be a consequence of the initial seeding or due to the test conditions having strong and persistent wind velocity from a single direction; that is, starting torque in low wind conditions is not a component of fitness in these experiments where twisted designs may be more beneficial.

\section{Conclusions}

Design mining represents a methodology through which the capabilities of computational intelligence can be exploited directly in the physical world through 3D printing. Our previous pilot study~\cite{Preen:2015} considered the parallel design of two interacting objects. In this paper we have used a well-known abstract model of coevolution to explore and extend various techniques from the literature, with an emphasis on reducing the number of fitness function evaluations. Our results suggest the round-robin evaluation process using the best solution in each other population is robust~\cite{Potter:1994}, as is our previously presented sampling method of surrogate models built using strictly population specific data. It has also been shown that the same techniques remain effective when scaling to 6 interacting species. These findings were than applied to a more complex version of the wind turbine design task considered in our pilot study, primarily moving from designing a pair of heterogeneous VAWT to a system of 6 turbines. As noted above, the SCGA remains robust to an increasing number of turbines since the number of inputs to the models remains constant. Indeed, we are unaware of a real-world coevolutionary optimisation task of this scale with or without surrogate models.

The VAWT spline representation used here is also much more flexible than the simple integer approach used previously, enabling the exploration of designs where the blades are not attached to the central shaft. This has enabled designs to emerge that not only exploit or compensate for the wind interaction with the central shaft, but also the effect of mass and vibrational forces as the turbines freely rotate around the mounted pins at high speed. That is, it has been shown possible to exploit the fan-generated wind conditions in the environment, including the complex inter-turbine turbulent flow conditions, and position-specific wind velocity, to design an array of 6 different turbines that work together collaboratively to produce the maximum angular kinetic energy. Note that the starting design for the turbines was based upon a standard commercial design and performance in the array was seen to double over the time allowed.

3D printing provides a very flexible way to rapid-prototype designs for testing. One of the most significant benefits of the technology is a `complexity paradox' wherein the complexity of the printed shape is highly uncorrelated with production time/cost. With conventional manufacturing, the more complex an object is, the more it costs to fabricate (especially when sub-components require complex assembly processes.) However, with 3D printing, the time and cost to fabricate an object mostly depends on the amount of material used. Moreover, the more complex a shape is, the more numerous the spaces (voids) that exist between components, and thus the smaller the quantity of material required. There is thus a synergy between computational intelligence techniques that can search a wide variety of complex shapes in a complex environment whilst also exploring the effects of novel materials. Here only PLA plastic was used to fabricate designs, however there are now over a hundred different materials that 3D printers can use, ranging from cells to titanium. Future work may explore the use of flexible materials and multi-material designs, which may result in very different designs of future wind farms. In addition, 3D printing can produce designs at different fidelity, such as slower more accurate prints for subtle optimisation or rapid coarse designs for quick evaluation. Fabrication can also be parallelised with multiple printers, e.g.,\ a different printer for each species.

Future work may also use the power generated as the objective under different wind conditions specific to the target environment, e.g., low cut-in speed; the design of larger wind farms, including turbine location, multiple rows of turbines, and collocation of VAWT and HAWT; the exploration of alternative surrogate models to reduce the number of fabrications required; alternative shape representations that can enable increased morphological freedom, including varying the number of blades, e.g., supershapes~\cite{Preen:2014}; and the use of novel fabrication materials. In addition, future avenues of research may include arrays of collaborating variable-speed wind turbines, turbines located on roof tops, and floating wind turbines. The use of adaptive design representations, allowing the number of shape parameters to increase as necessary (e.g.,~\cite{Olhofer:2001}), which will involve adaptive/coevolved surrogate models (e.g.,~\cite{Schmidt:2008}) will also be of interest. 

The issue of scalability remains an important area of future research. Changes in dimensionality may greatly affect performance, however it remains to be seen how the performance will change in the presence of other significant factors such as turbine wake interactions. Larger 3D printing and testing capabilities could be used to design larger turbines using the same method, although with longer fabrication and testing times. However, 3D printing is a rapidly developing technology capable of fabricating increasingly bigger parts with decreasing production times; for example, the EBAM~300 (Sciaky Ltd., USA) can produce a 10~ft long titanium aircraft structure in 48~hrs. On the micro-scale, turbines with a rotor diameter smaller than 2~mm can be used to generate power, e.g., for wireless sensors~\cite{Howey:2011}, and in this case high precision 3D printers would be required. Recently, 3D printing capabilities have been added to aerial robots to create flying 3D printers~\cite{Hunt:2014}, which may eventually enable swarms of robots to rapidly create, test, and optimise designs in areas that are difficult to access.    

The design mining approach outlined here provides a general and flexible framework for engineering design with applications that cannot be simulated due to the complexity of materials or environment. We anticipate that in the future, such approaches will be used to create highly unintuitive yet efficient bespoke solutions for a wide range of complex engineering design applications.
\clearpage

\clearpage
    
\section*{Acknowledgements}

This work was supported by the Engineering and Physical Sciences Research Council under Grant EP/N005740/1, and the Leverhulme Trust under Grant RPG-2013-344. The data used to generate the graphs are available at: \url{http://researchdata.uwe.ac.uk/166}.
\small
\bibliographystyle{apa-good}
\bibliography{full,references}
\end{document}